\documentclass[lettersize,journal]{IEEEtran}
\usepackage{amsmath,amsfonts}
\usepackage{algorithmic}
\usepackage{algorithm}
\usepackage{array}
\usepackage[caption=false,font=normalsize,labelfont=sf,textfont=sf]{subfig}
\usepackage{textcomp}
\usepackage{stfloats}
\usepackage{url}
\usepackage{verbatim}
\usepackage{graphicx}
\usepackage{cite}
\usepackage{multirow}

\usepackage[small]{caption}
\usepackage{booktabs}
\usepackage{makecell}
\usepackage{cleveref}
\usepackage{comment}
\usepackage{amssymb} 
\usepackage{subcaption}
\usepackage{multirow}
\usepackage{wrapfig}
\usepackage{color} 
\usepackage{xcolor} 
\usepackage{enumerate}
\usepackage{pifont}

\begin{document}

\title{Generalizing Unsupervised Lidar Odometry Model from Normal to Snowy Weather Conditions}

\author{Beibei Zhou, Zhiyuan Zhang, Zhenbo Song, Jianhui Guo, Hui Kong*
\thanks{Beibei Zhou is with Shanghai Polytechnic University, Shanghai, China (e-mail: beibeizhou18@gmail.com).
Zhiyuan Zhang is with Singapore Management University, Singapore (e-mail: cszyzhang@gmail.com).
Zhenbo Song and Jianhui Guo are with Nanjing University of Science and Technology, Nanjing, Jiangsu (\{songzb,guojianhui\}@njust.edu.cn).
Hui Kong is with University of Macau, Macau, China (e-mail: huikong@um.edu.mo). 
}
\thanks{*Correspondence author.}}

\markboth{Journal of \LaTeX\ Class Files,~Vol.~14, No.~8, August~2021}%
{Shell \MakeLowercase{\textit{et al.}}: A Sample Article Using IEEEtran.cls for IEEE Journals}


\maketitle

\begin{abstract}
Deep learning-based LiDAR odometry is crucial for autonomous driving and robotic navigation, yet its performance under adverse weather, especially snowfall, remains challenging. Existing models struggle to generalize across conditions due to sensitivity to snow-induced noise, limiting real-world use.  
In this work, we present an unsupervised LiDAR odometry model to close the gap between clear and snowy weather conditions. Our approach focuses on effective denoising to mitigate the impact of snowflake noise and outlier points on pose estimation, while also maintaining computational efficiency for real-time applications.
 To achieve this, we introduce a Patch Spatial Measure (PSM) module that evaluates the dispersion of points within each patch, enabling effective detection of sparse and discrete noise. 
 We further propose a Patch Point Weight Predictor (PPWP) to assign adaptive point-wise weights, enhancing their discriminative capacity within local regions. To support real-time performance, we first apply an intensity threshold mask to quickly suppress dense snowflake clusters near the LiDAR, and then perform multi-modal feature fusion to refine the point-wise weight prediction, improving overall robustness under adverse weather.
Our model is trained in clear weather conditions and rigorously tested across various scenarios, including snowy and dynamic. Extensive experimental results confirm the effectiveness of our method, demonstrating robust performance in both clear and snowy weather. This advancement enhances the model's generalizability and paves the way for more reliable autonomous systems capable of operating across a wider range of environmental conditions.
\end{abstract}


\section{Introduction}
\IEEEPARstart{L}{iDAR} 
sensors are indispensable in autonomous vehicles and robotics, providing precise 3D spatial data in point cloud form that enables critical tasks such as localization, mapping, and object detection~\cite{geiger2013vision, zhang2014loam}. Their ability to capture detailed environmental information across varying weather conditions is essential for the safe and reliable operation of autonomous systems~\cite{chen2015deepdriving}.
However, LiDAR sensors are vulnerable to noise, especially in adverse weather. Snowy conditions, in particular, introduce significant noise, which degrades 3D point cloud quality and affects the accuracy of ego-motion estimation—a fundamental aspect of autonomous navigation~\cite{zang2019impact}. Enhancing LiDAR odometry models to generalize effectively from normal to snowy weather conditions is essential for improving the dependability of autonomous systems in real-world, varied environments.
\label{sec:intro}
 \begin{figure} [t] 
		\centering  
		\includegraphics[width=0.95\linewidth]{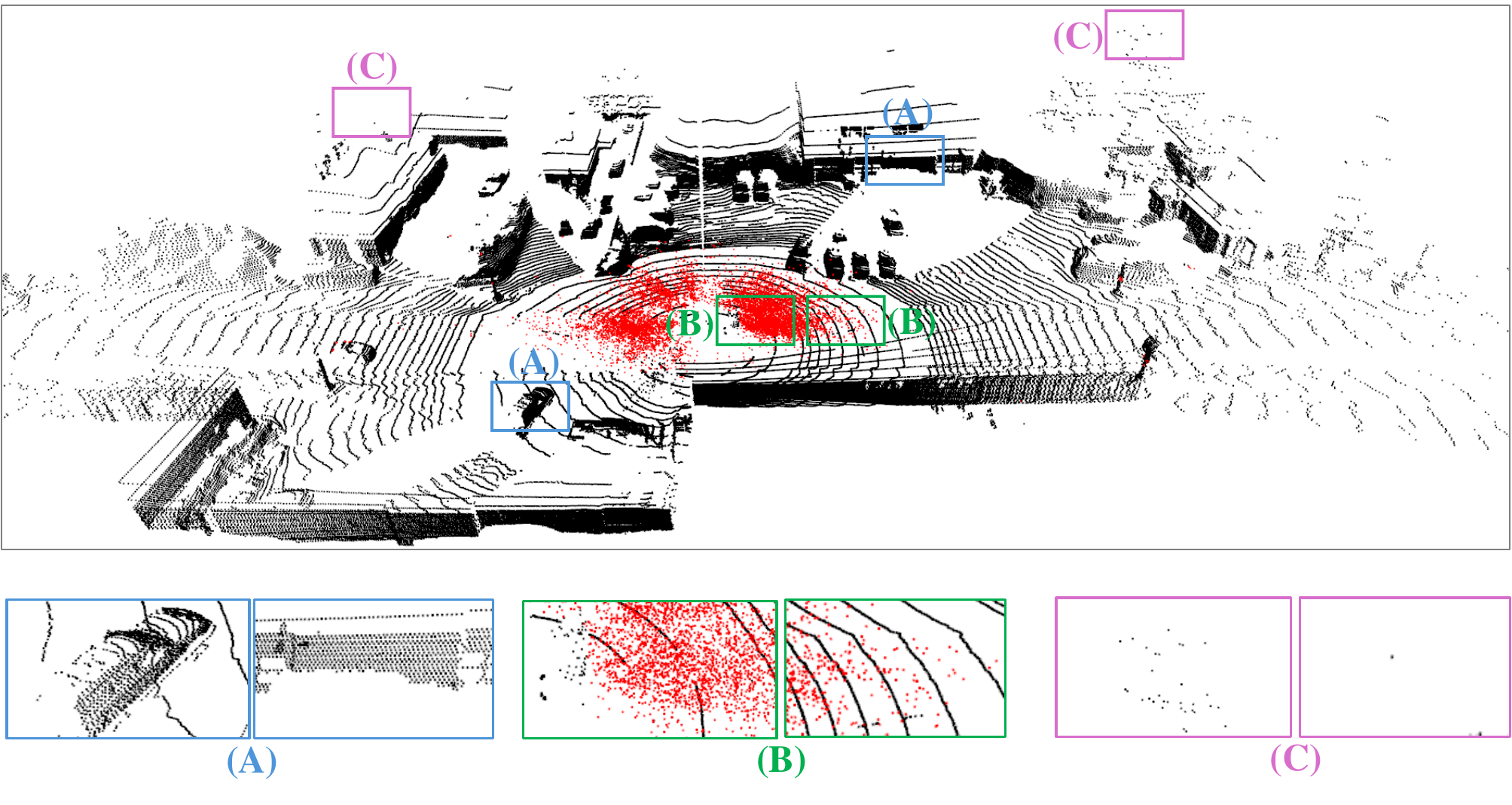}  
  \caption{Visualization of a point cloud  under snowfall conditions: (A) scene points, (B) snowflake noise points, (C) sparsely distributed and isolated points at a distance. }
		\label{pc}  
		\vspace{-0.3cm}
\end{figure}

Traditional statistical denoising methods, such as ROR and SOR~\cite{carrilho2018statistical}, often assume overly simplified noise patterns, resulting in the loss of crucial environmental details and reducing accuracy in real-world applications. While dynamic adaptations like DROR~\cite{charron2018noising}, DSOR~\cite{kurup2021dsor}, and DIOR~\cite{roriz2021dior} improve feature preservation through adaptive thresholds, they still struggle with computational inefficiency due to extensive neighbor-point searches. These limitations hinder their integration with modern deep-learning-based odometry networks.
Recent deep learning approaches, such as WeatherNet~\cite{heinzler2020cnn}, 4DenoiseNet~\cite{seppanen20224denoisenet}, and LiSnowNet~\cite{yu2022lisnownet}, have shown promise in denoising under adverse weather. 
However, these models are highly dependent on large, high-quality training datasets, which limits their effectiveness in unfamiliar environments. Their primary focus is on denoising tasks, and similar to traditional approaches, they require substantial computational resources and training time for denoising. Integrating such a model into our system would significantly reduce the speed of the odometry.
To overcome these challenges, we propose an unsupervised LiDAR odometry model that generalizes robustly from clear to snowy conditions. 
It incorporates a denoising module and a hierarchical structure, enabling accurate localization across diverse environments without the need for ground truth labels.

Our model is built upon three key observations: (1) Point clouds are denser near the LiDAR sensor and become increasingly sparse with distance, making correspondence matching more challenging; (2) Snowflake points typically exhibit lower intensity values compared to genuine scene points; and (3) Snow tends to form dense clusters in close proximity to the sensor, while distant snowflakes appear more dispersed (\Cref{pc}).
Guided by these insights, we propose two novel modules: the Patch Spatial Measure (PSM) and the Patch Point Weight Predictor (PPWP). 
PSM partitions the point cloud into local patches and computes spatial autocorrelation scores to quantify the density of each patch, with higher scores indicating denser areas. 
This approach identifies the distant sparse scene points and snowflakes described in \Cref{pc}(B) and (C), and assigns them lower spatial correlation scores to reduce their negative impact on pose estimation caused by missing accurate correspondences.
PPWP is a two-stage module designed for fine-grained denoising by assessing point-wise confidence within each patch, in contrast to PSM which evaluates dispersion at the patch level.
The first stage, Intensity Threshold Masking, rapidly filters out low-intensity points in high-density corresponding to near-field snowflakes. The second stage employs a Multi-Modal Point-wise Weight Predictor (MPWP), which integrates point feature, global feature, intensity, and distance via a cross-attention fusion mechanism. This design improves the reliability of weight prediction by adaptively assigning confidence scores to individual points based on multi-modal information.
Finally, benefiting from the model's emphasis on scene structure comprehension and low sensitivity to snowflake noise (which is effectively suppressed through Intensity Threshold Masking), our model can directly generalize from clear weather conditions to snowy scenarios with similar structural layouts while remaining largely unaffected by snowfall interference.

Our main contributions are as follows:
\begin{itemize}
\item We present an end-to-end unsupervised LiDAR odometry model that operates effectively in both normal and snowy conditions, eliminating the need for labeled data.
 
\item We develop a Patch Spatial Measure (PSM) to calculate spatial auto-correlation scores, which reduces the impact of sparse and isolated noise points on pose estimation.

\item We introduce a Patch Point Weight Predictor (PPWP) with a threshold mask and multi-modal fusion, which effectively remove the gathered snowflake points and discern the significance of each point within a patch.

\item Our method demonstrates robust generalization and superior pose accuracy across the KITTI Odometry, Ford, and WADS datasets.
\end{itemize}

\section{Related Work}
\label{Related Work}

\subsection{Deep LiDAR Odometry} 

Deep learning-based LiDAR odometry (LO) has evolved from supervised to more advanced unsupervised approaches. Early supervised methods, such as CNN-based LO~\cite{velas2018cnn}, LO-Net~\cite{li2019net}, DeepPCO~\cite{wang2019deeppco} and Translo~\cite{translo}, focused on predicting poses by projecting 3D point clouds into 2D images to facilitate keypoint matching, which increased computational demands. Notably, Velas et al.~\cite{velas2018cnn} framed the LO task as a classification problem, while LO-Net~\cite{li2019net} enhanced the network's capabilities by incorporating normal estimation and mask prediction modules. 
DeepPCO~\cite{wang2019deeppco} employed a deeply parallel neural network architecture to estimate translation and rotation separately, yielding accurate results at the cost of high computation. 
Translo applies a masked point transformer framework for robust scene estimation.
Recent advancements like PWCLO-Net~\cite{wang2021pwclo}, EfficientLO-Net~\cite{wang2022efficient}, and DELO~\cite{Ali_2023_ICCV} avoid 2D projections by directly processing 3D point clouds, yielding better efficiency and accuracy. However, their performance is limited to specific datasets.

In contrast, unsupervised LO methods have gained popularity for their flexibility and lower data requirements, especially valuable when labeled data is scarce or costly. Cho et al.~\cite{cho2020unsupervised} pioneered unsupervised LO by leveraging geometry-aware losses from point-to-plane ICP~\cite{low2004Linear} as a supervisory signal. Subsequent works, such as SelfVoxeLO~\cite{SelfVoxeLO}, introduced new architectures for self-supervised learning: SelfVoxeLO uses 3D voxel features with multiple loss functions.
To further enhance pose estimation, recent models like HPPLO-Net~\cite{zhou2023hpplo} and Li et al.~\cite{li20233d} utilize a coarse-to-fine structure aided by scene flow, improving both accuracy and computational efficiency. Overall, unsupervised LO approaches are becoming increasingly valuable for autonomous systems, as they provide robust and adaptable solutions without the reliance on extensive labeled datasets.

\subsection{Point Cloud Denoising}

Early statistical methods such as Radius Outlier Removal (ROR) and Statistical Outlier Removal (SOR) \cite{carrilho2018statistical} used fixed thresholds based on distance or the number of neighboring points to detect and remove outliers. These methods have evolved into dynamic approaches like Dynamic Radius Outlier Removal (DROR) \cite{charron2018noising} and Dynamic Statistical Outlier Removal (DSOR) \cite{kurup2021dsor}, designed specifically for autonomous driving. DROR, for instance, identifies outliers by dynamically adjusting the search radius based on neighboring points, while DSOR calculates mean and variance in distances among a fixed set of nearest neighbors to improve adaptability. LIOR \cite{park2020fast} extends these methods by filtering low-intensity points, which are more likely to represent noise. Advanced methods~\cite{roriz2021dior, wang2022scalable} incorporate both distance and intensity metrics, but they still require extensive neighbor searches, which pose challenges for real-time systems.

Deep learning based methods have accelerated the denoising techniques, beginning with PointNet \cite{qi2017pointnet} and extending to point cloud-specific networks like PCN \cite{rakotosaona2020pointcleannet}. WeatherNet \cite{heinzler2020cnn} leverages semantic segmentation models adapted for adverse weather, while recent self-supervised methods \cite{bae2022slide, yu2022lisnownet} have advanced snow noise removal without requiring labeled data. LiSnowNet \cite{yu2022lisnownet} uses discrete small wavelet and inverse discrete wavelet transforms for denoising, and 4DenoiseNet \cite{seppanen20224denoisenet} incorporates time-sequential data with semantic segmentation for robust noise classification. Despite their potential, deep learning-based methods remain dependent on high-quality, diverse datasets for optimal performance.

\begin{figure*}[t]
	\centering
		\includegraphics[width=0.95\linewidth]{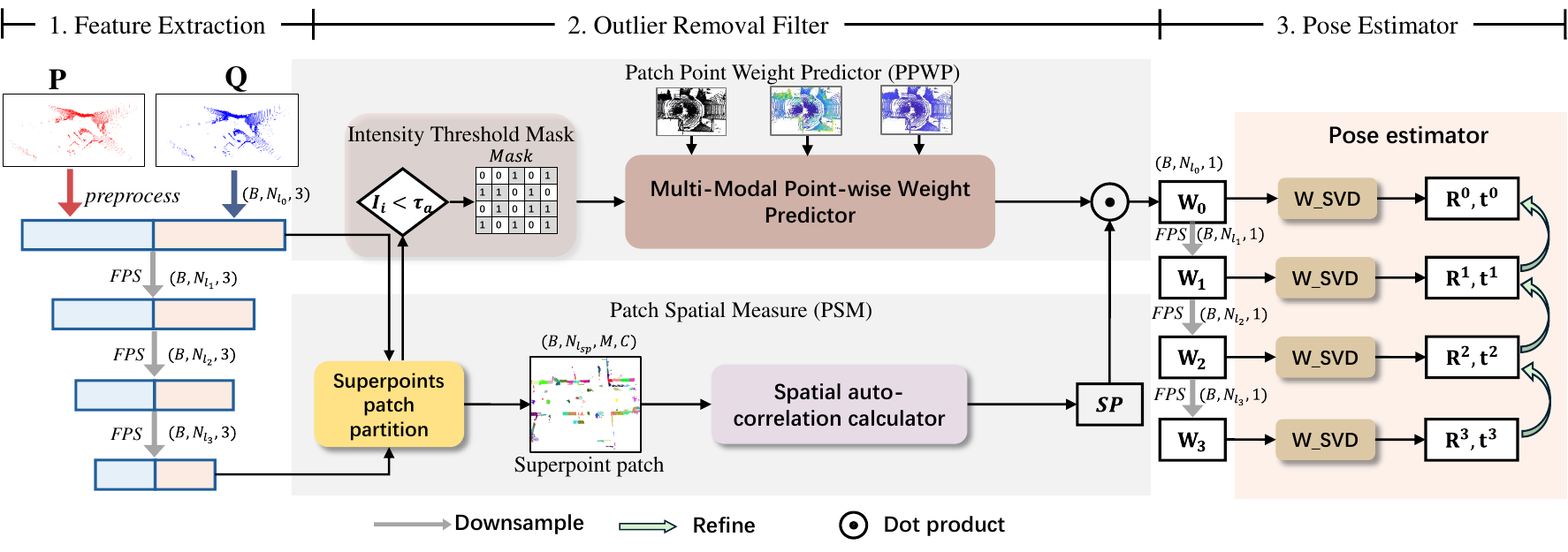}	
	\caption{Overview of our framework: 1. Feature Extraction extract multi-scale features from the two frames $P$ and $Q$ and the points at the coarsest layer serve as the superpoints. 2. Outlier Removal Filter uses the superpoints to segment the finest layer point cloud into patches. For each patch, two advanced modules PSM and PPWP are designed to quantify the point dispersion and evaluate the point-wise confidence. 3. Pose Estimator performs multi-scale point matching between the two frames and estimate the relative pose.
	}
	\label{overview}
	\vspace{-0.3cm}
\end{figure*}

\section{Method}
\label{sec:method}

The framework of our method is shown in~\Cref{overview} which comprises three components: Feature Extraction, Outlier Removal Filter, and Pose Estimator.

\subsection{Feature extraction}
To extract informative features, we develop a multi-scale feature pyramid within our network. The process begins with voxel downsampling, reducing the original point cloud to $N_{l_0}$ points as the input to the network, forming the highest resolution layer ($l_0$). Next, we apply Farthest Point Sampling (FPS)~\cite{qi2017pointnet++} to iteratively reduce the point set from $l_0$ to progressively lower resolution layers $l_1$, $l_2$, and $l_3$, with each layer containing one-fourth the points of the previous. This hierarchical structure enables efficient multi-scale feature extraction.

We then use a non-local network to extract features from the top layer $l_0$, downsampling these layers via PointConv~\cite{wu2019pointconv} to populate the feature sets of subsequent layers. This multi-scale module can also optimize pose estimation in a coarse-to-fine approach, using the initial coarse layer estimate to reduce computational load (\Cref{sec:pose_estimator}). The feature set for each layer is denoted as $F_l \in \mathbb{R}^{N_{l} \times C}$.

\subsection{Outlier removal filter}
The Outlier Removal Filter is a key component of our system, comprising two advanced modules: Patch Spatial Measure (PSM) and Patch Point Weight Predictor (PPWP). 
The PSM module is designed to compute a spatial auto-correlation score by analyzing the dispersion of points within a localized region to distinguish scattered noises and others, 
while the PPWP evaluates the point-wise confidence value of each point within the patch via multi-modal information.

\begin{figure*}[t]
\centering
\includegraphics[width=1\textwidth]{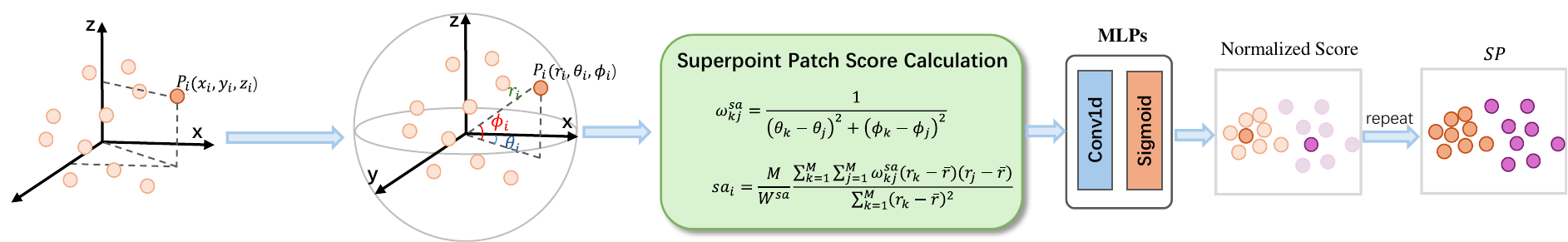} 
\caption{Patch Spatial Measure (PSM) Module.}
\label{PSM}
\end{figure*}

 \subsubsection{Patch Spatial Measure}
 
 This module begins by segmenting the $l_0$ layer point cloud into multiple patches based on superpoints. We select points at the coarsest layer $l_3$ of $P$ as superpoints and identify the closest $M$ points surrounding each superpoint on layer $l_0$ to construct a local patch. Consequently, the finest layer $l_0$ is divided into $N_{l_3}$ patches, represented as:  
\vspace{-0.1cm}
 \begin{equation}
 \small
 \mathcal{G}^P = \{g_i \in \mathbb{R}^{M \times 3} | i=1,2,\dots N_{l_3} \}.  \label{patch}
\end{equation}

Following the approach in~\cite{zhang2023detecting}, we introduce spatial auto-correlation serves as a valuable metric for evaluating the dispersion of point clouds by measuring each point's correlation with its neighbors. 
We compute the discretization of each patch by integrating spatial auto-correlation score and MLP centered on superpoint according to the characteristics of our task. The specific details is shown in~\Cref{PSM}.
As the final output of this module, a higher score indicates a tightly clustered patch, suggesting it is integral to the scene. Conversely, a lower score implies more dispersed points, potentially indicating scattered snowflakes or distant sparse elements, which can hinder two-frame alignment and adversely affect pose estimation.

Since the $l_0$ layer is obtained through voxel downsampling of the raw point cloud, it roughly preserves the spatial structure and dispersion characteristics of the point cloud. So it is rational to calculate the spatial auto-correlation on the $l_0$ layer.  
 Given a set of point clouds:
\vspace{-0.2cm}
 \begin{equation}
 \small
P=\{p_{i}=(x_i, y_i, z_i)| i=1,2,\dots N_{l_0}\},
\end{equation}
we first transform the point cloud coordinates from the Cartesian coordinate system to the spherical coordinate system, which allows us to more intuitively obtain the distance and angular information for each point. The coordinate transformation process is as follows:
\vspace{-0.2cm}
\begin{align}
 \small
& r_i = \sqrt{x_i^2 + y_i^2 + z_i^2}, \\
& \theta_i = \arctan\left(\frac{y_i}{x_i}\right), 
\phi_i = \arcsin\left(\frac{z_i}{r_i}\right). 
\end{align}
Then the point cloud $P$ can be represented as~\Cref{spherical}:
\vspace{-0.2cm}
\begin{equation}
\small
P=\{p_{i}=(r_i, \theta_i, \phi_i)| i=1,2,\dots N_{l_0}\}, 
\vspace{-0.2cm}\label{spherical}
\end{equation}
where \( r_i \),  \( \theta_i \) and \( \phi_i \) represent the distance, azimuth and elevation respectively.

We apply Global Moran's I~\cite{moran1950notes}\cite{li2007beyond} to determine the spatial auto-correlation $SA = \{sa_i | i=1,2,\dots, N_{l_3}\}$ for each patch $g_i$, as depicted in~\Cref{sa}. This captures the spatial affinity between points, considering both their distances and angular relationships.
\vspace{-0.2cm}
\begin{equation}
\small
    sa_i= \frac{M}{W^{sa}}
\frac{\sum_{k=1}^{M}\sum_{j=1}^{M} w_{kj}^{sa}(r_{k}-\bar{r})(r_{j}-\bar{r})}{\sum_{k=1}^{M}(r_{k}-\bar{r})^{2}},  \label{sa}
\vspace{-0.2cm}
\end{equation}
where $\bar{r}=\frac{1}{M}\sum_{k=1}^{M}r_{k}$ is the mean distance of the set of points in $g_i$.
The weight $w_{kj}^{sa}$ captures the angular discrepancy between a point and its neighbors, where greater discrepancies correspond to diminished weight values, as delineated in~\Cref{wij}:
\vspace{-0.2cm}
\begin{equation}
\small
w_{kj}^{sa} = \frac{1}{(\theta_k - \theta_j)^2 + (\phi_k - \phi_j)^2}, \label{wij}
\end{equation}
and $W^{sa}=\sum_{k=1}^{M}\sum_{j=1}^{M}w_{kj}^{sa}$ is the sum of $w_{kj}^{sa}$.
In our computation of spatial autocorrelation, we factor in both the Euclidean distances and angular separations of points, enabling a more nuanced discrimination of spatial structures. 
We use \( \theta \) and \( \phi \) to calculate the angular deviation of neighboring points from point $k$. The inverse of the deviation is used as a weight value for each point pair, so when the absolute distances are the same, that equal absolute distances are weighted more heavily for smaller angular deviations, resulting in a higher spatial auto-correlation score.

The values of $SA \in \mathbb{R}^{N_{sp}\times1}$, ranging from $[-1,1]$, are fed into MLPs for normalization to [0, 1] as shown in~\Cref{sp}. Then the normalized scores are allocated to every point within their respective patches, signifying the local dispersion for each point. Compared to direct linear normalization of $SA$, MLPs can conduct complex non-linear mappings tailored to specific data distributions through training, enabling adaptive responses to variations in the scene.
Then the spatial autocorrelation score ($SP$) of the superpoint is extended to every point in the patch, corresponding to $repeat$, as shown in~\Cref{sp}:
\begin{equation}
SP = repeat(MLP(SA)) \in \mathbb{R}^{N_{l_0}\times1}.  
\label{sp}
\end{equation}

\begin{figure}[t]
\centering
\includegraphics[width=0.48\textwidth]{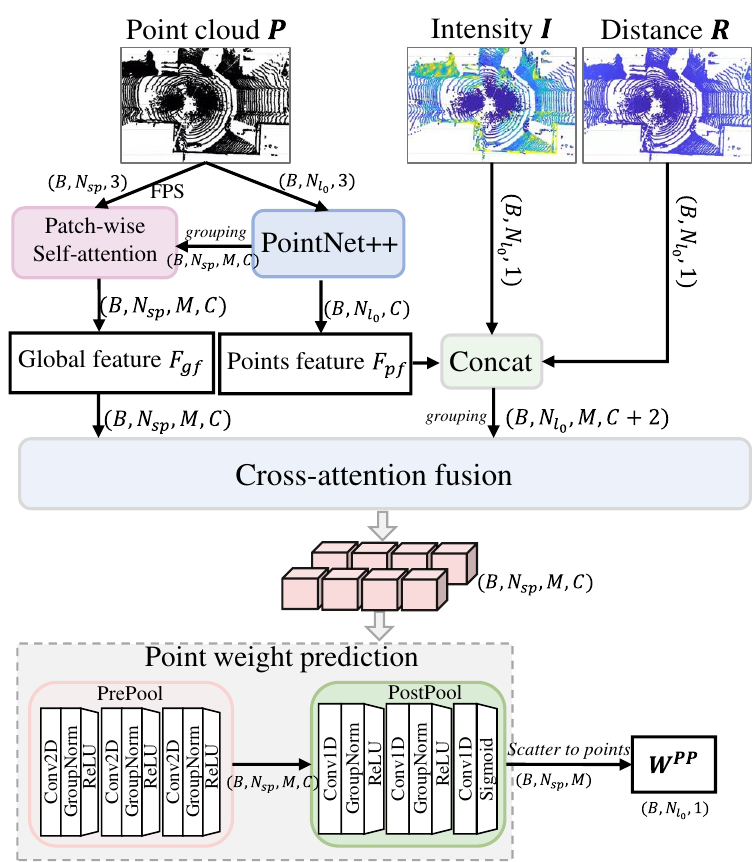} 
\caption{
\textbf{Multi-modal Point-wise Weight Predictor.} It takes the point feature, global feature, intensity, and distance as input, integrating them via a cross-attention fusion mechanism. The fused multi-modal feature is then used to predict point-wise weights via the Point Weight Predictor.
}
\vspace{-0.4cm}
\label{MMPWP}
\end{figure}

\subsubsection{Patch Point Weight Predictor}
 
This module consists of an intensity threshold mask and a multi-modal point-wise weight predictor (MPWP). The intensity threshold mask is a simple concept, but together with the MPWP, it can swiftly and effectively suppress the negative impact of snowflake noise.

\paragraph{{Intensity Threshold Mask}} 
To preliminarily filter out clustered snowflake noise near the LiDAR, we apply an intensity threshold mask ($M$) before predicting point-wise weights.
Instead of relying on statistical methods\cite{carrilho2018statistical,qi2017pointnet} or convolutional networks\cite{bae2022slide, yu2022lisnownet}, we adopt a empirical intensity threshold ($\tau_a$) to improve efficiency, better aligning with the real-time requirements of our pose estimation task.
Due to the strong diffuse reflection and multi-directional scattering caused by snowflakes, the return signal strength is significantly reduced compared to other objects in the scene. Therefore, we set an empirical intensity threshold of $\tau_a = \text{Max}(I) * 0.01$ to initially filter out snowflake points in the scene.
Points that fall beneath this threshold are flagged as snowflake points and are designated with the mask $m_i$ set to 0, as outlined in~\Cref{mask}.
\begin{align}
\small
\vspace{-0.2cm}
 M=\{m_i|i=1,2,\dots N_{l_0}\}, \\  
 m_i = 
\begin{cases} 
0, & \text{if } I_i < \tau_a \\
1, & \text{otherwise}.
\vspace{-0.2cm}
\label{mask}
\end{cases}
\end{align}
where $I=\{I_{i}| i=1,2,\dots N_{l_0}\}$ represents the intensity value of the point cloud.

\paragraph{{Multi-modal Point-wise Weight Predictor}} 
To accurately estimate the point-wise confidence within a patch, we incorporate multi-modal information, including the point intensity $I$, distance $R$, point-wise feature $\mathbf{F}_{pf}$, and patch-level global feature $\mathbf{F}_{gf}$. The detailed architecture of the proposed MPWP module is illustrated in Figure~\ref{MMPWP}. 
The point feature $\mathbf{F}_{pf}$ is extracted via PointNet++~\cite{qi2017pointnet++} and grouped into patches, forming $\mathbf{F}_{2} \in \mathbb{R}^{(B, N_{sp}, M, C)}$, where $N_{sp}$ and $M$ represent the number of superpoints and points per patch, respectively.

We employ a self-attention mechanism to extract local contextual features for each patch. The grouped feature $\mathbf{F}_{2} \in \mathbb{R}^{(B, N_{sp}, M, C)}$ is projected into query, key, and value representations as follows:
\begin{equation}
\mathbf{Q}_2 = \mathbf{F}_{2} W_{Q_2}, \quad \mathbf{K}_2 = \mathbf{F}_{2} W_{K_2}, \quad \mathbf{V}_2 = \mathbf{F}_{2} W_{V_2},
\end{equation}
where $W_{Q_2}, W_{K_2}, W_{V_2} \in \mathbb{R}^{C \times d_k}$ are learnable parameters, and $\mathbf{Q}_2, \mathbf{K}_2, \mathbf{V}_2 \in \mathbb{R}^{(B, N_{sp}, M, d_k)}$.
The self-attention map $A_2$ is calculated as follows:
\begin{equation}
\small
\vspace{-0.2cm}
A_2 = \text{softmax} \left( \frac{\mathbf{Q}_2 \mathbf{K}_2^\top}{\sqrt{d_k}} \right) \in \mathbb{R}^{(B, N_{sp}, M, M)},
\end{equation}
where softmax normalizes the pairwise relationships across the $M$ points within each patch.
The global feature ${F}_{gf}$ is aggregated via:
\begin{equation}
\small
\vspace{-0.2cm}
\mathbf{F}_{\text{attn}} = A_2 \cdot \mathbf{V}_2 \in \mathbb{R}^{(B, N_{sp}, M, d_k)}.
\end{equation}
\begin{equation}
\mathbf{F}_{gf} = \mathbf{F}_{\text{attn}} W_O \in \mathbb{R}^{(B, N_{sp}, M, C)}, \quad W_O \in \mathbb{R}^{d_k \times C},
\end{equation}
the output $\mathbf{F}_{gf}$ provides patch-wise local features that capture fine-grained geometric relationships, facilitating downstream point-wise confidence estimation. Here, $W_O$ is a learnable linear projection matrix that maps the attention output features from dimension $d_k$ back to the original feature dimension $C$.

To enhance point-wise confidence estimation reliability, we apply a multi-modal cross-attention fusion module, integrating ${F}_{gf}$, $F_{pf}$, $I$ and $R$ into a unified feature space to generate fine-grained point-wise weights $W^{pp} \in \mathbb{R}^{(B, N_{l_0}, 1)}$. We first concatenate $F_{pf}$, $I$, and $R$ to form a fused input, grouping it into patches $\mathbf{F}_{group} \in \mathbb{R}^{(B, N_{sp}, M, C+2)}$ using the same FPS grouping indices from $\mathbf{F}_{group}$. For each patch, we treat $\mathbf{F}_{gf}$ as the query, and $\mathbf{F}_{group}$ as key and value, respectively.
\begin{equation}
\mathbf{Q} = \mathbf{F}_{gf} W_Q, \quad \mathbf{K} = \mathbf{F}_{group} W_K, \quad \mathbf{V} = \mathbf{F}_{group} W_V,
\end{equation}
where $W_Q \in \mathbb{R}^{C \times d_k}$, $W_K, W_V \in \mathbb{R}^{(C+2) \times d_k}$, and $\mathbf{Q}, \mathbf{K}, \mathbf{V} \in \mathbb{R}^{(B, N_{sp}, M, d_k)}$.
The cross-attention map is computed as:
\begin{equation}
\small
A = \text{softmax} \left( \frac{\mathbf{Q} \cdot \mathbf{K}^T}{\sqrt{d_k}} \right) \in \mathbb{R}^{(B, N_{sp}, M, M)}.
\end{equation}
Then the fused features are obtained by:
\begin{equation}
\small
\mathbf{F}_{\text{fused}} = A \cdot \mathbf{V} \in \mathbb{R}^{(B, N_{sp}, M, d_k)}.
\end{equation}
$\mathbf{F}_{\text{fused}}$ combines patch-level context with point-wise multi-modal information, yielding more informative and adaptive point features.

To predict the final point-wise weights, we adopt a two-stage structure comprising $PrePool$ and $PostPool$. $PrePool$ leverages Conv2D for joint patch-level and local feature aggregation, whereas $PostPool$ utilizes Conv1D to refine intra-patch point features and predict precise point-wise weights. 
The fused features are then passed through this module as~\Cref{weight}:
\begin{equation}
W_{pp'} = PostPool(PrePool(\mathbf{F}_{\text{fused}})) \in \mathbb{R}^{(B, N_{sp}, M)}. \label{weight}
\end{equation}
Finally, we scatter $W^{pp'}$ back to the original point cloud space and resolve any conflicts (e.g., select maximum or first-come strategy) to get the weight of each point
$W^{pp} = \{w_i^{pp}\} \in \mathbb{R}^{(B, N_{l_0}, 1)}.$

The final weight $W$ of each point is the element-wise product of the spatial autocorrelation score $SP$, the updated mask $M^{update}$, and the patch point weight $W^{pp}$ as shown in~\Cref{gf2}:
 \begin{equation}
 \small
  W = SP\circ M\circ W^{pp} \label{gf2}.
\end{equation}

The weight matrix $W$ is derived through multi-scale Farthest Point Sampling (FPS), aligning with the scales of the point clouds $P_{l_1}$, $P_{l_2}$, and $P_{l_3}$, respectively. This approach bolsters computational efficiency by enabling a single computation at level $l_0$ to ascertain the weights for all subsequent layers. Rather than incorporating a dedicated module for snowflake point discrimination, our network architecture capitalizes on threshold filtering, grounded in numerical analysis, to expeditiously eliminate snowflakes. Remaining snowflakes are treated as typical noise, which can be mitigated based on the scores and confidence levels output by PSM and MPWP. This integrated strategy ensures that our model adeptly navigates through snowy conditions without compromising the integrity of the pose estimation.

\subsection{Pose estimator}
\label{sec:pose_estimator}

As depicted in~\Cref{overview}, we employ point matching to establish point correspondences by estimating the scene flow between the source point cloud at each level, akin to the methodology in PointPWC~\cite{wu2020pointpwc}. Initially, we determine the coarsest pose at level $l_3$:
\begin{equation}
\small
    D_{l_3} = P_{l_3} + SF_{l_3}  \label{warpflow}.
\end{equation}
\begin{equation}
\small
    [\textbf{R}^3, \textbf{t}^3] = W\_po2pl\_SVD(W_{l_3}, P_{l_3}, D_{l_3}, Nor_{l_3}),  \label{svd}
\end{equation}
where $SF_{l_3}$ is the scene flow between the source point cloud $P_{l_3}$ and the target point cloud $Q_{l_3}$, which is employed to synthesize a pseudo-target point cloud $D_{l_3}$. Then we use the weighted point-to-plane SVD (denoted $W\_po2pl\_SVD$) proposed in \cite{zhou2023hpplo} to decompose the pose matrix by~\Cref{svd}, where $\mathbf{R}^3\in\mathbb{R}^{3\times 3}$ and $\mathbf{t}^3\in\mathbb{R}^{3\times 1}$ represent the rotation matrix and translation vector, respectively.
With the initial pose matrix at level $3$, we apply the transformation to the point cloud at level $l_2$ as \Cref{svd2}:
\begin{equation}
\small
    P_{l_2}^{warp} = {\mathbf{R}^3}P_{l_2}+\mathbf{t}^3, \label{svd2}
\end{equation}
 and repeat the process of~\Cref{warpflow} to~\Cref{svd2} between $P_{l_2}^{warp}$ and $Q_{l_2}$ at level $2$. This yields the residual pose matrix at level $l_2$: $[\Delta \mathbf{R}^2, \Delta \mathbf{t}^2]$. We then update the pose at level $l_2$ as follows:
\begin{equation}	
\small
\mathbf{R}^2 = \Delta\mathbf{R}^2\cdot{\mathbf{R}^3}, \\
\end{equation}
\begin{equation}	
\small
\mathbf{t}^2 = \Delta\mathbf{R}^2\cdot{\mathbf{t}^3} + \Delta{\mathbf{t}^2}.
\end{equation}
Similarly, we continue updating the pose matrix at levels $1$ and $0$, and can obtain the final pose matrix at level $l_0$: $[\mathbf{R}^{0}, \mathbf{t}^0]$. This matrix is the result of three stages of optimization across levels $l_2$, $l_1$, and $l_0$, representing the network's final pose output.

\subsection{Loss Function}

Our approach employs a point-to-plane Iterative Closest Point (ICP) loss function at each layer to refine the estimated pose, as detailed in~\Cref{loss}. This method ensures precise point correspondences by focusing on the densest subsets of both the source point cloud $P$ and the target point cloud $Q$. We proceed by transforming $P$ and then determining the nearest neighbor in $Q$ to establish these correspondences. The point-to-plane loss for these matched point pairs is derived as follows:
{
\begin{multline}
    loss = \{\sum_{l=0}^{3}\sum_{{p}\in{P}}{M}^l({\mathbf{R}^l} \cdot p + \mathbf{t}^l - q) \cdot n_q) \, | \\
    \min_{q} {(\mathbf{R}^l} \cdot p + \mathbf{t}^l - q)^2, 
    q \in Q, n_q \in {Nor_Q}\},  \label{loss}
\end{multline}
}
where $n_q \in {Nor_Q}$ represents the normals of $Q$.

\section{Experiment}
\label{sec:result}
Our model is trained exclusively on the KITTI Odometry sequences, and its generalization capabilities are rigorously evaluated across three distinct datasets: the KITTI Odometry Dataset under normal weather, the Ford Campus Vision and Lidar Dataset (Ford) with numerous dynamic objects, and the Winter Adverse Driving Dataset (WADS), which captures the complexities of snowy conditions. This comprehensive testing strategy enables us to assess the model's robustness and versatility in handling a wide range of real-world scenarios, from standard to extreme environments.

\subsection{Data preprocessing}

We first computed the normal vectors for each point in the original point cloud, with parameters set to \(radius = 4\) and \(max_{nn} = 50\). Given the high density of the original point cloud, the normals derived from it are considered to be reliable.
Subsequently, we employed voxel downsampling to regulate the scale of the input point cloud at layer \(l_0\) to 8192 points, with each voxel cell measuring 0.2 meters on each side. This method effectively maintains the spatial structure and the sparse distribution of the original point cloud, which is crucial for the subsequent calculation of spatial correlation scores.
Also, we eliminated ground points within 0.5 meters from the ground to focus on the relevant scene elements. Points beyond 30 meters from the vehicle were also excluded due to their sparsity and the challenges they present in matching across consecutive frames. 
These preprocessing steps improve data quality and make our model more robust in different driving conditions.

\subsection{Experimental Setting}
        \begin{table}[h]	
		\centering
    		\caption{Hardware and software environment used for the model.}
		\resizebox{1\linewidth}{!}{
			\begin{tabular}{|l|l|} \toprule[1.5pt]
				{Hardware environment}  & {Software environment}  \\ \hline
				{Processor cores: 20 Core}  &Ubuntu 18.04  \\
				{RAM: 31G} &CUDA11.8, CuDNN8.9.2.26   \\
				{GPU: NVIDIA GeForce RTX 3090Ti} &Python 3.9.17   \\ 
				{Dedicated GPU memory: 24G}  &Pytorch 2.0.1 \\ \toprule[1.5pt]
		\end{tabular}}
		\label{tab:environment}
		\vspace{-0.3cm}	
	\end{table}

Our model is developed using PyTorch on a single NVIDIA GeForce RTX 3090 GPU and optimized using the Adam optimizer. We configure the hyperparameters with values of \(\beta_1=0.9\), \(\beta_2=0.99\), and \(W_{decay}=10^{-5}\). 
The learning rate is initialized to 0.001, with a step decay scheduler that decrements the rate every 20 epochs, a strategy that was implemented on the KITTI Odometry training dataset~\cite{geiger2012we}.
The specific software and hardware configurations employed in our approach are detailed in Table~\ref{tab:environment}.

\begin{table}[t] 
		\setlength\tabcolsep{2pt}    
		\centering
	\caption{Evaluation results on the KITTI Odometry test set. $t_{rel}$: average translational root mean square error (RMSE) drift (\%); $r_{rel}$: average rotational RMSE drift $(^{\circ}/100m)$. The best performance in each group is highlighted in bold.}	
		\resizebox{1\linewidth}{!}{
			\begin{tabular}{l||ll|ll|ll|ll|ll}
				\toprule[1.5pt]
	
                    \multicolumn{1}{c||}{\multirow{2}*{Methods}} 
    
				& \multicolumn{2}{c|}{07}  & \multicolumn{2}{c|}{08}   &\multicolumn{2}{c|}{09}
				& \multicolumn{2}{c|}{10}
				& \multicolumn{2}{c}{Avg.07-10} \\ \cline{2-11}  
				\multicolumn{1}{c||}{~} & {$t_{rel}$}   & {$r_{rel}$}  & $t_{rel}$  & $r_{rel}$
				& {$t_{rel}$}   & {$r_{rel}$}  & $t_{rel}$  & $r_{rel}$
				& {$t_{rel}$}   & {$r_{rel}$}  \\ \toprule[1pt]

Full A-LOAM~\cite{aloam}   &{{0.69}}&{{0.50}} &{\textbf{1.18}}&{\textbf{0.44}} &{\textbf{1.20}}&{\textbf{0.48}} &{1.51}&{\textbf{0.57}}  &{\textbf{1.15}}&{\textbf{0.50}}  \\  \cline{2-11} 

A-LOAM\cite{aloam}  &2.89 &1.80 &4.82 &2.08 &5.76 &1.85 &3.61 &1.76 &3.89 &1.64 \\ 

ICP-po2po~\cite{ICP}  &5.17 &3.35 &10.04 &4.93 &6.93 &2.89 &8.91 &4.74 &7.76 &3.98 \\

ICP-po2pl~\cite{ICP}  &1.55 &1.42 &4.42 &2.14 &3.95 &1.71 &6.13 &2.60 &4.01 &1.97  \\  

VGICP\cite{koide2021voxelized}   &{\textbf{0.64}} &{\textbf{0.45}} &1.58 &0.75 &1.97 &0.77 &\textbf{1.31} &{0.62} &1.38 &0.65 \\  

CLS\cite{velas2016collar}  &1.04 &0.73 &2.14 &1.05 &1.95 &0.92 &3.46 &1.28 &2.15 &1.00 \\  \toprule[1pt]

LO-Net~\cite{li2019net}   &1.70 &0.89 &2.12 &0.77 &{1.37} &{0.58} &1.80 &0.93 &1.75 &0.79  \\ 	

PWCLO-Net~\cite{wang2021pwclo}  &0.60&0.44&\textbf{1.26}&\textbf{0.55}&\textbf{0.79}&\textbf{0.35}&{1.69}&\textbf{0.62}   &{\textbf{1.09}}&\textbf{0.49} \\
DELO~\cite{ali2023delo}  &\textbf{0.58}&\textbf{0.41}&1.36&0.64&1.23&0.57&\textbf{1.53}&0.90  &1.18&0.63   \\   \toprule[1pt]

Cho et al.~\cite{cho2020unsupervised}   &-&- &-&-  &4.87&1.95 &5.02&1.83 &4.95&1.89\\

Delora\cite{nubert2021self}  &-&-   &-&-  &6.05&2.15  &6.44&3.00  &6.25&2.58 \\

SelfVoxelLO~\cite{SelfVoxeLO}  &3.09&1.81 &3.16&1.14 &3.01&1.14 &3.48&1.11 &3.19&1.30  \\
            
RSLO~\cite{xu2022robust} 
&2.37&1.15 &2.14&0.92 &2.61&1.05 &2.33&0.94 &2.36&1.02       \\
HPPLO-Net~\cite{zhou2023hpplo}   &{0.89}&{0.57}  &\textbf{1.12}&\textbf{0.54}  &1.45&{0.61}  &{1.22}&{\textbf{0.66}}   &{1.17}&{0.59}      \\  \cline{2-11} 
            
Ours  &\textbf{0.50}&\textbf{0.40}&1.30&0.60&\textbf{1.16}&\textbf{0.61}&\textbf{1.21}&0.68   &\textbf{1.04}&\textbf{0.57}  \\
    		
				\toprule[1pt]
    \end{tabular}}		    
		\label{tab:seq0010}
\end{table}
\vspace{-0.5cm}
\begin{table}[t]
    \centering
    \caption{The Evaluation results on Ford test set. $t_{rel}$: average translational root mean square error (RMSE) drift (\%); $r_{rel}$: average rotational RMSE drift $(^{\circ}/100m)$. The best performance in each group is highlighted in bold.}
    \resizebox{0.95\linewidth}{!}{
			\begin{tabular}{c|c||cc|cc}
				\toprule[1.3pt]
        \multicolumn{2}{c||}{\multirow{2}*{Methods}} 
        & \multicolumn{2}{c|}{Ford-1} & \multicolumn{2}{c}{Ford-2} \\ \cline{3-6} 
        \multicolumn{2}{c||}{~}& $t_{rel}$ & $r_{rel}$ & $t_{rel}$ & $r_{rel}$ \\   \toprule[1.3pt]

        \multirow{6}{*}{Traditional}
				&{Full A-LOAM~\cite{aloam}}  &\textbf{1.88}&\textbf{0.50} &\textbf{2.05}&\textbf{0.56}  \\  \cline{2-6}
    &A-LOAM   &4.17 &2.00 &{4.72} &1.65   \\ 
    &{ICP-po2po} &8.20 &2.64  &16.2 &2.84    \\ 
    &{ICP-po2pl}   &3.35 &1.65  &5.68 &1.96    \\ 
    &{VGICP}~\cite{koide2021voxelized}     &{\textbf{2.51}} &{\textbf{1.03}}  &{\textbf{3.65}} &{\textbf{1.39}}  \\
    &CLS~\cite{velas2016collar}  &10.5 &3.90    &14.7 &4.60   \\ \toprule[1pt]
	\multirow{2}{*}{Supervised}			
				&{LO-Net~\cite{li2019net}}   &\textbf{2.27} &\textbf{0.62}  &\textbf{2.18} &\textbf{0.59}   \\
    &{PWCLO-Net}~\cite{wang2021pwclo} &5.89 &2.02  &16.84  &4.62 \\ 
    \toprule[1pt]
    \multirow{4}{*}{Unsupervised}	
    &RSLO~\cite{xu2022robust}  &19.14 &6.55 &16.74&4.55 \\ 
    &{HPPLO-Net}~\cite{zhou2023hpplo} &2.52 &\textbf{1.82} &3.48 &1.47 \\\cline{2-6}
    &{Ours} &\textbf{2.45}&1.96  &\textbf{3.46}&\textbf{1.25} \\

				\toprule[1.3pt]
    \end{tabular}}
		\label{tb:ford}
		\vspace{-0.5cm} 
\end{table}

\begin{table*}[t] 
		\setlength\tabcolsep{5pt}    
      \caption{Evaluation results on the WADS dataset. $t_{rel}$: average translational root mean square error (RMSE) drift (\%); $r_{rel}$: average rotational RMSE drift $(^{\circ}/100m)$. The best performance in each group is highlighted in bold.}
\centering

\resizebox{1\linewidth}{!}{
    \begin{tabular}{l||ll|ll|ll|ll|ll|ll|ll|ll|ll}   
        \toprule[1.5pt]
\multirow{2}{*}{Methods} 
& \multicolumn{2}{c|}{11}  & \multicolumn{2}{c|}{12}   &\multicolumn{2}{c|}{13}
& \multicolumn{2}{c|}{14} & \multicolumn{2}{c|}{15} & \multicolumn{2}{c|}{16}  & \multicolumn{2}{c|}{17}   &\multicolumn{2}{c|}{18}
& \multicolumn{2}{c}{20} \\ \cline{2-19}  
 & {$t_{rel}$}   & {$r_{rel}$}  & $t_{rel}$  & $r_{rel}$& {$t_{rel}$}   & {$r_{rel}$}
& {$t_{rel}$}   & {$r_{rel}$}  & $t_{rel}$  & $r_{rel}$& {$t_{rel}$}   & {$r_{rel}$}
& {$t_{rel}$}   & {$r_{rel}$} & {$t_{rel}$}   & {$r_{rel}$} & {$t_{rel}$}   & {$r_{rel}$} \\ \toprule[1pt]
			
    Full A-LOAM &0.92&\textbf{0.49}&4.59&0.77&0.96&0.51&\textbf{0.99}&\textbf{0.90}&1.80&0.92&1.16&\textbf{0.83}&7.33&0.97&2.48&0.62&1.72&0.82 \\ \cline{2-19}
    
    A-LOAM &8.25&1.47&17.12&2.08&14.09&2.20&5.44&1.02&13.39&0.93&6.94&2.06&15.67&1.03&22.30&5.86&13.55 &1.46\\ 
    
    ICP-po2po &\textbf{0.73}&0.74&{2.41}&{1.18}&1.70&1.51&1.64&2.01&2.70&1.21&1.48&1.93&7.68&2.03&2.45&1.42&2.04 &1.22\\
    
    ICP-po2pl  &1.91&1.50&2.28&1.72&\textbf{0.83}&\textbf{0.31}&1.72&1.99&\textbf{1.80}&\textbf{0.91}&\textbf{0.97}&1.07&7.12&1.19&1.47&0.68&\textbf{0.75} &\textbf{0.70} \\  
    
    VGICP &1.17&0.85&\textbf{1.71}&\textbf{0.69}&1.21&0.76&1.01&1.15&1.86&1.30&1.27&1.10&\textbf{6.62}&\textbf{0.77}&\textbf{1.47}&\textbf{0.32}&1.73&0.96\\  
\toprule[1pt]

    PWCLO-Net &{12.85}&{17.14}&{62.19}&{99.19}&{10.71}&{17.76}&{56.71}&{50.84}&{25.47}&{14.83}&{21.16}&{18.38}&{96.25}&{115.44}&{21.15}&{16.37}&{55.45}&{122.88} \\ \hline
        
    RSLO &46.08&8.95 &58.41&6.40 &75.05&11.10 &67.82&14.66 &67.59&9.45 &97.60&28.67 &73.11&52.90 &72.17&14.51 &53.23&5.94 \\
    HPPLO-Net &\textbf{0.97}&1.02&\textbf{1.56}&\textbf{0.63}&\textbf{0.80}&0.58&0.76&0.72&1.61&0.66&1.46&1.33&6.80&\textbf{1.68}&1.43&\textbf{0.50}&2.06&\textbf{0.82} \\  \cline{2-19}
				
    Ours &1.43&\textbf{0.90}&2.00&0.69&0.89&\textbf{0.34}&\textbf{0.18}&\textbf{0.42}&\textbf{1.20}&\textbf{0.51}&\textbf{1.06}&\textbf{1.18}&\textbf{6.39}&{1.87}&\textbf{0.99}&0.71&\textbf{1.02}&{0.93} \\     		
				\toprule[1.5pt]
\multirow{2}{*}{Methods}   & \multicolumn{2}{c|}{22} & \multicolumn{2}{c|}{23}  & \multicolumn{2}{c|}{28} & \multicolumn{2}{c|}{30}  & \multicolumn{2}{c|}{34} & \multicolumn{2}{c|}{35} & \multicolumn{2}{c|}{36} & \multicolumn{2}{c|}{37} & \multicolumn{2}{c}{avg}  \\ \cline{2-19}  
 & {$t_{rel}$}   & {$r_{rel}$}  & $t_{rel}$  & $r_{rel}$& {$t_{rel}$}   & {$r_{rel}$}
& {$t_{rel}$}   & {$r_{rel}$}  & $t_{rel}$  & $r_{rel}$& {$t_{rel}$}   & {$r_{rel}$}
& {$t_{rel}$}   & {$r_{rel}$} & {$t_{rel}$}   & {$r_{rel}$} & {$t_{rel}$}   & {$r_{rel}$} \\ \toprule[1pt]
			
Full A-LOAM &\textbf{1.10}&\textbf{0.36}&3.21&0.49&\textbf{0.80}&\textbf{0.55}&4.16&0.98&1.05&\textbf{0.43}&1.16&\textbf{0.43}&\textbf{0.74}&\textbf{0.51}&0.84&0.85&2.06&0.68\\  \cline{2-19}

A-LOAM &9.13&1.53&17.01&1.99&11.28&2.56&13.01&3.90&3.70&0.61&5.51&0.81&9.42&2.13&11.90&4.99&11.63&2.15\\ 

ICP-po2po &3.36&0.58&11.52&0.96&1.99&0.52&5.59&1.44&1.11&0.62&1.27&0.85&0.92&0.90&0.73&0.74&2.90&11.17\\

ICP-po2pl &1.45&1.27&3.52&2.47&1.27&0.96&4.15&0.73&2.11&1.36&\textbf{1.08}&0.99&0.79&0.98&\textbf{0.52}&\textbf{0.72}&1.99&1.15 \\  

VGICP &1.45&0.71&\textbf{1.07}&\textbf{0.78}&0.99&0.65&\textbf{3.90}&\textbf{1.29}&\textbf{1.04}&0.98&1.20&0.89&0.78&0.57&0.77&1.13&\textbf{1.72}&\textbf{0.88} \\  \toprule[1pt]

PWCLO-Net &{61.20}&{12.03}&{59.58}&{26.63}&{23.87}&{29.43}&{21.15}&{34.57}&{20.09}&{65.94}&{7.48}&{157.76}&{86.90}&{134.78}&{45.58}&{67.12}&{42.81}&{58.88}\\ \hline
    
RSLO &136.76&74.19 &78.47&14.63 &83.58&13.17 &88.19&16.36 &49.66&11.66 &47.83&15.93 &53.38&15.42 &46.07&13.88 &70.29&19.28 \\
HPPLO-Net &4.19&6.91&6.56&\textbf{3.45}&1.22 &0.41&3.92&1.40&1.30&\textbf{0.62}&1.27&\textbf{0.39}&\textbf{0.63}&\textbf{0.58}&0.85&\textbf{0.47}&2.20&1.30  \\  \cline{2-19}
            
Ours &\textbf{1.09}&\textbf{2.27}&\textbf{2.97}&5.13 &\textbf{0.68}&\textbf{0.25}&\textbf{3.22}&\textbf{0.66}&\textbf{1.14}&0.77&\textbf{0.86}&0.61&0.64&0.61&\textbf{0.49}&0.81&\textbf{1.54}&\textbf{1.10} \\     		
				\toprule[1pt]
    \end{tabular}}		
		\label{tab:wads}
\end{table*}

\subsection{Evaluation results on KITTI Odometry}

Our model was trained on sequences 00-06 of the KITTI Odometry dataset and subsequently tested on sequences 07-10. We compared our approach with a comprehensive set of methods, including six traditional methods: Full A-LOAM~\cite{aloam}, A-LOAM~\cite{aloam}, ICP-po2po~\cite{ICP}, ICP-po2pl~\cite{ICP}, VGICP~\cite{koide2021voxelized}, and CLS~\cite{velas2016collar}; three supervised methods: LO-Net~\cite{li2019net}, PWCLO-Net~\cite{wang2021pwclo}, and DELO~\cite{ali2023delo}; and five unsupervised methods: Cho et al.~\cite{cho2020unsupervised}, DeLORA~\cite{nubert2021self}, SelfVoxelLO~\cite{SelfVoxeLO}, RSLO~\cite{xu2022robust}, and HPPLO-Net~\cite{zhou2023hpplo}. 
The evaluation results for these methods are presented in Table~\ref{tab:seq0010}. We utilized the ICP-po2po and ICP-po2pl versions from PyICP-SLAM~\cite{ICP}, which are renowned for enhancing performance and robustness in SLAM systems, and have shown promising results across various datasets. To ensure a fair comparison, the loop closure detection module was disabled.
Among the traditional methods, full A-LOAM, which incorporates back-end optimization, demonstrated the highest performance. The supervised method PWCLO-Net is acknowledged as one of the top performers among deep learning-based methods under clear weather. Despite lacking back-end optimization and being trained unsupervised, our method performs comparably and achieves the lowest overall average translational error, highlighting its effectiveness.
The strong performance of our method can be attributed to several key factors.
Firstly, our model endeavors to retain as many robust points as feasible, thereby enhancing the overall reliability of the point cloud data. Secondly, it mitigates the detrimental impact of noise on the pose matrix decomposition, a common challenge in odometry tasks. Lastly, our approach optimizes the pose estimation in a layer-by-layer manner, which incrementally enhances the accuracy of the estimated trajectory. This meticulous optimization allows our method to outperform counterparts, providing more accurate and stable odometry estimation.

\subsection{The Generalization Evaluation on Ford}
\label{evaluation_Ford}

To assess the generalizability of our model, we extended our evaluation to the Ford Dataset, renowned for its longer driving distances, intricate environmental settings, and a substantial presence of dynamic objects. These characteristics render the Ford dataset an exemplary benchmark for gauging the model's adaptability across diverse scenarios. Our model, trained on the KITTI Odometry sequences 00-06, was subjected to testing on Ford-1 and Ford-2 sequences.
A comparative result of our method with existing approaches is presented in Table \ref{tb:ford}. The performance metrics were derived from the results published by the authors in their papers or through their publicly accessible code repositories. Our method surpasses all unsupervised methods and the majority of traditional methods. Notably, it falls just short of Full A-LOAM, which incorporates a sophisticated back-end optimization module, thereby highlighting the robustness and effectiveness of our unsupervised approach in the absence of such enhancements.
It is noteworthy that PWCLO-Net and RSLO, which heavily rely on convolutional layers to predict intricate 6-DoF poses, exhibit diminished generalization capabilities when evaluated across diverse datasets. This is particularly evident in the WADS dataset, where their performance is significantly challenged.
In contrast, our method employs convolutional networks exclusively for feature extraction and point correspondence identification, leveraging the Patch Spatial Measure (PSM) module to account for local geometric structures. Subsequently, we utilize SVD to decompose the pose matrix, thereby reducing reliance on convolutional neural networks. This strategic approach allows each module to operate within its domain of expertise, enhancing the overall performance of our model. Consequently, our model demonstrates resilience and robustness even in more intricate dynamic scenarios, as evidenced by its consistent performance across the Ford dataset.

\subsection{The Generalization Evaluation on WADS}

To assess the generalizability of our method under snowy conditions, we conducted tests on the Winter Adverse Driving Dataset (WADS) without requiring retraining. This dataset has a scene structure similar to that of KITTI but features different weather conditions. We selected 17 sequences, each exceeding 100 meters in length, and compared our method with eight state-of-the-art LiDAR odometry algorithms, as detailed in Table~\ref{tab:wads}.
These methods were evaluated on WADS using the open-source code provided by the authors, and the parameters are set to the optimal values recommended by the authors. Notably, the deep learning-based models, including PWCLO-Net, RSLO, HPPLO-Net, and ours, were uniformly trained on sequences 00-06 of the KITTI Odometry dataset to maintain fairness in the evaluation process.
The table illustrates that our method outperforms all other methods in terms of the overall average error, highlighting its exceptional performance and robustness even in the challenging snowy conditions of the WADS dataset.

Particularly, deep learning-based methods experience a notable decline in performance when transitioning from normal weather to snowy conditions, which introduce substantial noise. In stark contrast, our approach retains its efficacy in snowy environments, primarily attributed to two key factors: firstly, the network architecture and pose estimation strategy as elucidated in~\Cref{evaluation_Ford}; 
and secondly, the independence of our method from network-based snowflake point removal, opting instead for a swift filtering of dense snowflake points via an efficient mask threshold.
Within the network, the Patch Spatial Measure (PSM) and Patch Point Weight Predictor (PPWP) modules are tasked with processing the remaining discrete noise particles, which are analogous to the sparse noise typically encountered in clear weather conditions. As a result, our model perceives no significant discrepancy between clear and snowy scenarios. This methodology not only conserves considerable time in snowflake removal but also bolsters the model's generalizability across diverse environmental conditions.
This dual-pronged approach ensures that our model is not only efficient in adverse weather but also robust against the variability inherent in real-world applications, thereby setting a new standard for LiDAR odometry in autonomous systems.

\vspace{-0.2cm}
\subsection{Ablation study}
        \begin{table}[t] 
		\setlength\tabcolsep{3pt}   
  		\caption{The ablation study results of different modules on KITTI odometry. $t_{rel}$: average translational root mean square error (RMSE) drift (\%); $r_{rel}$: average rotational RMSE drift $(^{\circ}/100m)$. The best performance in each group is highlighted in bold.}	
		\centering
		
		\resizebox{1\linewidth}{!}{
			\begin{tabular}{l||ll|ll|ll|ll|ll}
				\toprule[1.5pt]
\multirow{2}{*}{Methods} 
& \multicolumn{2}{c|}{07}  & \multicolumn{2}{c|}{08}   &\multicolumn{2}{c|}{09}
& \multicolumn{2}{c|}{10}
& \multicolumn{2}{c}{Avg.07-10} \\  \cline{2-11}  
& {$t_{rel}$}   & {$r_{rel}$}  & $t_{rel}$  & $r_{rel}$
& {$t_{rel}$}   & {$r_{rel}$}  & $t_{rel}$  & $r_{rel}$
& {$t_{rel}$}   & {$r_{rel}$}  \\ \toprule[1pt]
Baseline Method  &{1.19}&{0.74}  &{1.77}&{0.69 } &{1.71}&{0.72}  &{1.73}&{0.99}  &{1.60}&{0.78 } \\ \hline
PSM &\underline{0.65}&0.55&\underline{1.28}&0.61&\textbf{0.99}&0.52&1.57&0.82 &1.12&0.63 \\ 
MPWP  &0.87&\underline{0.51}&1.37&\underline{0.60}&1.11&\textbf{0.49}&\underline{1.06}&\underline{0.56} &1.10&\textbf{0.54} \\
PSM+MPWP  &0.95&0.64&\textbf{1.23}&0.68&\underline{1.05}&\underline{0.50}&\textbf{0.90}&\textbf{0.56} &\textbf{1.03}&0.59 \\  
 PSM+MPWP+$M$ &\textbf{0.50}&\textbf{0.40}&1.30&\textbf{0.60}&{1.16}&{0.61}&{1.21}&0.68   &\underline{1.04}&\underline{0.57} \\  
				\toprule[1pt]
    \end{tabular}}	
		\label{ab:kitti}
		\vspace{-0.3cm}
\end{table}

\begin{table}[t] 
    \setlength\tabcolsep{4pt}    
     \caption{The ablation study results of different module on WADS.$t_{rel}$: average translational root mean square error (RMSE) drift (\%); $r_{rel}$: average rotational RMSE drift $(^{\circ}/100m)$. The best performance in each group is highlighted in bold.}
\centering
\resizebox{1\linewidth}{!}{
    \begin{tabular}{c||cc|cc|cc|cc|cc}
        \hline
        \multirow{2}{*}{Seq}   & \multicolumn{2}{c|}{\makecell[c]{Baseline\\}}     & \multicolumn{2}{c|}{\makecell[c]{PSM}}       & \multicolumn{2}{c|}{\makecell[c]{MPWP}} &\multicolumn{2}{c|}{\makecell[c]{PSM+MPWP}}
        & \multicolumn{2}{c}{\makecell[c]{PSM+MPWP+$M$}}
  \\ \cline{2-11} 
            & {$t_{rel}$}   & {$r_{rel}$}  & $t_{rel}$  & $r_{rel}$
        &$t_{rel}$   & $r_{rel}$  & $t_{rel}$  &$r_{rel}$
        &$t_{rel}$   & $r_{rel}$     \\ \hline
			
11  &\underline{0.94}&1.23 &1.46&0.90 &\textbf{0.87}&0.94 &{0.94}&\textbf{0.78}
   &1.43&\underline{0.90}     \\

12  &\textbf{1.37}&\textbf{0.54} &1.89&{0.60}  &\underline{1.46}&0.82 &1.60&\underline{0.59}
 &2.00&0.69   \\

13   &0.95&\underline{0.40} &\underline{0.90}&0.74&1.01&0.43 &1.29&0.55   &\textbf{0.89}&\textbf{0.34}  \\

14   &0.93&0.87 &\underline{0.42}&0.97 &0.71&0.87 &0.69&\underline{0.78}   &\textbf{0.18}&\textbf{0.42}  \\

15     &1.50&0.79 &\textbf{1.02}&0.98 &1.44&\underline{0.51}  &1.31&0.84
   &\underline{1.20}&\textbf{0.51}   \\

16   &1.72&1.38 &\textbf{0.92}&1.46 &1.75&1.52 &1.71&1.63  &\underline{1.06}&\textbf{1.18} \\

17 &7.00&1.51 &6.72&1.71 &6.99&\underline{1.51}  &\underline{6.69}&\textbf{1.46} &\textbf{6.39}&1.87 \\

18  &1.46&0.60 &1.01&\underline{0.38} &1.20&\textbf{0.29} &\textbf{0.85}&0.45  &\underline{0.99}&0.71 \\

20 &1.46&\textbf{0.85}  &\underline{1.18}&1.04 &1.91&0.94 &1.86&1.18  &\textbf{1.02}&\underline{0.93} \\

22    &3.05&5.54 &3.05&4.02 &\underline{2.04}&4.71 &\textbf{0.99}&6.50  &\underline{1.09}&\textbf{2.27} \\
23   &7.94&8.88 &3.29&5.58 &\underline{2.42}&\textbf{2.20} &\textbf{2.37}&5.50   &2.97&\underline{5.13}  \\
28   &1.14&0.60 &\textbf{0.39}&0.27 &0.95&\textbf{0.21} &\underline{0.43}&0.40 &0.68&\underline{0.25} \\
30    &3.92&1.41 &\underline{3.36}&0.78 &4.16&0.87 &3.70&\textbf{0.60}  &\textbf{3.22}&\underline{0.66} \\
34    &1.33&0.89 &1.15&0.72 &\textbf{1.11}&\textbf{0.60} &1.19&\underline{0.67}  &\underline{1.14}&0.77 \\
35    &1.37&0.84 &\textbf{0.72}&0.76 &\underline{0.79}&\textbf{0.34} &0.80&\underline{0.42} &0.86&0.61 \\
36   &\underline{0.70}&\textbf{0.53} &0.81&0.62 &0.74&\underline{0.56} &0.83&0.59  &\textbf{0.64}&0.61 \\
37    &0.97&0.85 &\textbf{0.28}&\textbf{0.62} &0.72&\underline{0.68} &0.60&0.90  &\underline{0.49}&0.81 \\  \hline

avg   &2.22&1.63 &1.68&1.30  &1.78&\textbf{1.06} &\underline{1.64}&1.40  &\textbf{1.54}&\underline{1.10} \\  
\toprule[1pt]	
 \end{tabular}}
	\label{ab:wads}
	\vspace{-0.3cm}
\end{table}

This section evaluates the efficacy of each module—the Patch Spatial Measure (PSM) and the Patch Point Weight Predictor (PPWP), which consists of the Multimodal Point-Wide Weight Predictor (MPWP) and the Intensity Threshold Mask (\(M\)), under varying weather conditions.

\paragraph{The effectiveness of each module under normal weather} 
We conducted tests on sequences 07-10 of the KITTI Odometry dataset to discretely assess the performance of each module. The experimental outcomes are detailed in Table~\ref{ab:kitti}. Our baseline approach is constituted by the fundamental framework devoid of three modules (Patch Spatial Measure (PSM), Multi-modal Point-wise Weight Predictor (MPWP), Intensity Threshold Mask (M)). As delineated in the table, Compared to the baseline method, the PSM and MPWP modules exhibit significant improvements. Moreover, under clear weather conditions, the inclusion of $M$ doesn't negatively affect the final results (see the comparison between PSM+MPWP and PSM+MPWP+$M$ in the table).
The incorporation of PSM and MPWP individually precipitates a marked reduction in errors; however, their combined implementation (PSM+MPWP) yields only a modest enhancement in performance.
This outcome can be attributed to the diminished noise levels in clear weather conditions, where the predominant challenges are the establishment of correspondences for sparse points and the handling of scattered noise. Both PSM and MPWP address these issues by apportioning weights to the point clouds, thereby effectively alleviating the difficulties encountered under normal weather conditions. 
Threshold masks ($M$) show limited error reduction, as they are specifically designed for snowy scenarios and have negligible impact on typical non-snowy scenes.
Essentially, our findings highlight the complementary roles of PSM and MPWP in enhancing the robustness of LiDAR odometry within local regions under clear weather conditions. In contrast, the threshold mask is optimized for heavy snowfall scenarios, while the removal of a small number of low-threshold points in clear weather has negligible impact on the final results.

\begin{figure*} [t] 
		\centering  
		\includegraphics[width=0.9\linewidth]{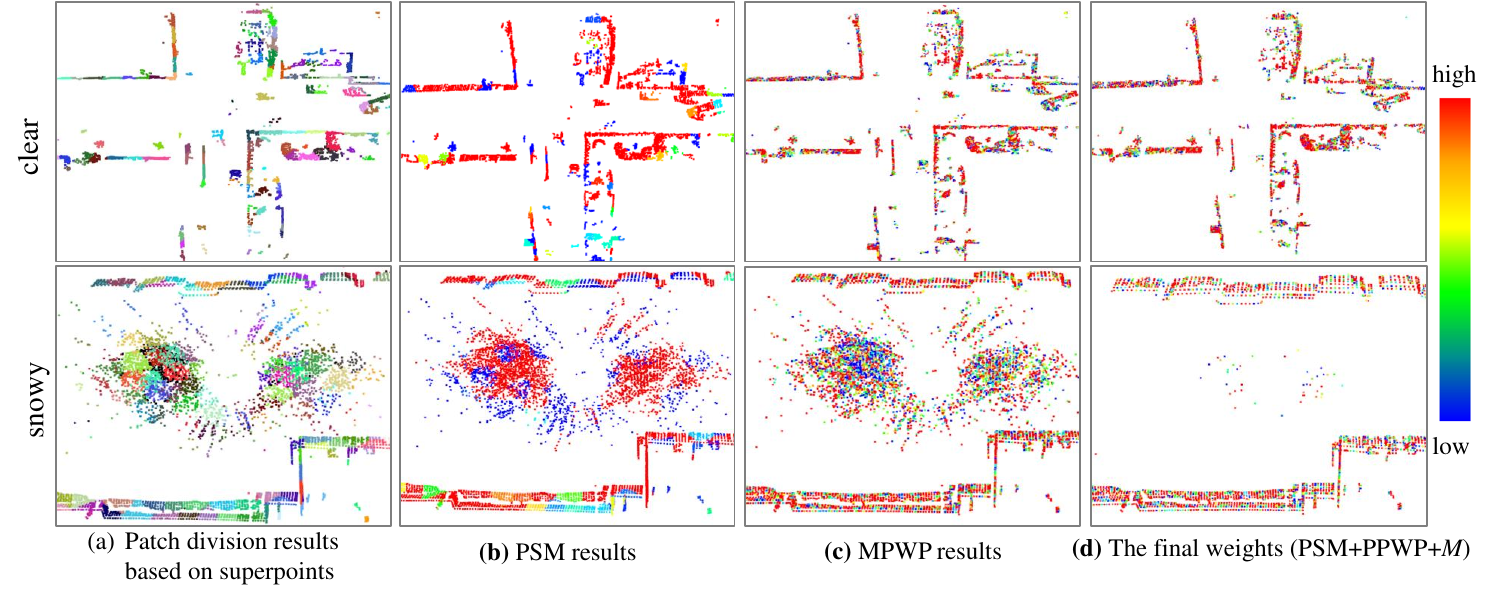}  
		\caption{Visualization of each module on clear and snowy weather. (a) Superpoint segmentation results, different colors represent different patches. (b) PSM assigns lower scores to dispersed snowflake points, higher scores to dense clusters near the sensor. (c) MPWP adaptively lowers confidence for most snowflake points, compensating for PSM’s limitations. (d) With the mask, most snowflake points are filtered out.}  
		\label{fig:wads_weight_viz}  
		\vspace{-0.2cm}
\end{figure*}

\paragraph{The effectiveness of each module on snowy weather.} 
To substantiate the efficacy of each module under snowy conditions, an ablation study was conducted across 17 sequences of the WADS dataset. The outcomes are detailed in Table~\ref{ab:wads}. The results clearly indicate that the incorporation of both the PSM and MPWP modules substantially bolsters the network's performance. 
A comparison between PSM+MPWP and PSM+MPWP+$M$ reveals that the threshold mask $M$ exerts a significant influence, leading to a pronounced decrease in both average translational and rotational errors.
Using the threshold mask, the majority of snowflakes are effectively eliminated, while the remaining residual noise is assigned minimal weights through PSM and MPWP, significantly improving pose estimation accuracy. This improvement is reflected in the results of PSM+MPWP+$M$ in Table~\ref{ab:wads}. Visualization examples for each module are displayed in Figure~\ref{fig:wads_weight_viz}. 
These findings underscore the pivotal role of the PSM, MPWP, and $M$ modules in mitigating the adverse effects of snowflake noise on LiDAR odometry, thereby ensuring robustness in snowy environments.

\vspace{-0.2cm}
\section{Visualization}

 \begin{figure}[t]
\centering
\includegraphics[width=0.48\textwidth]{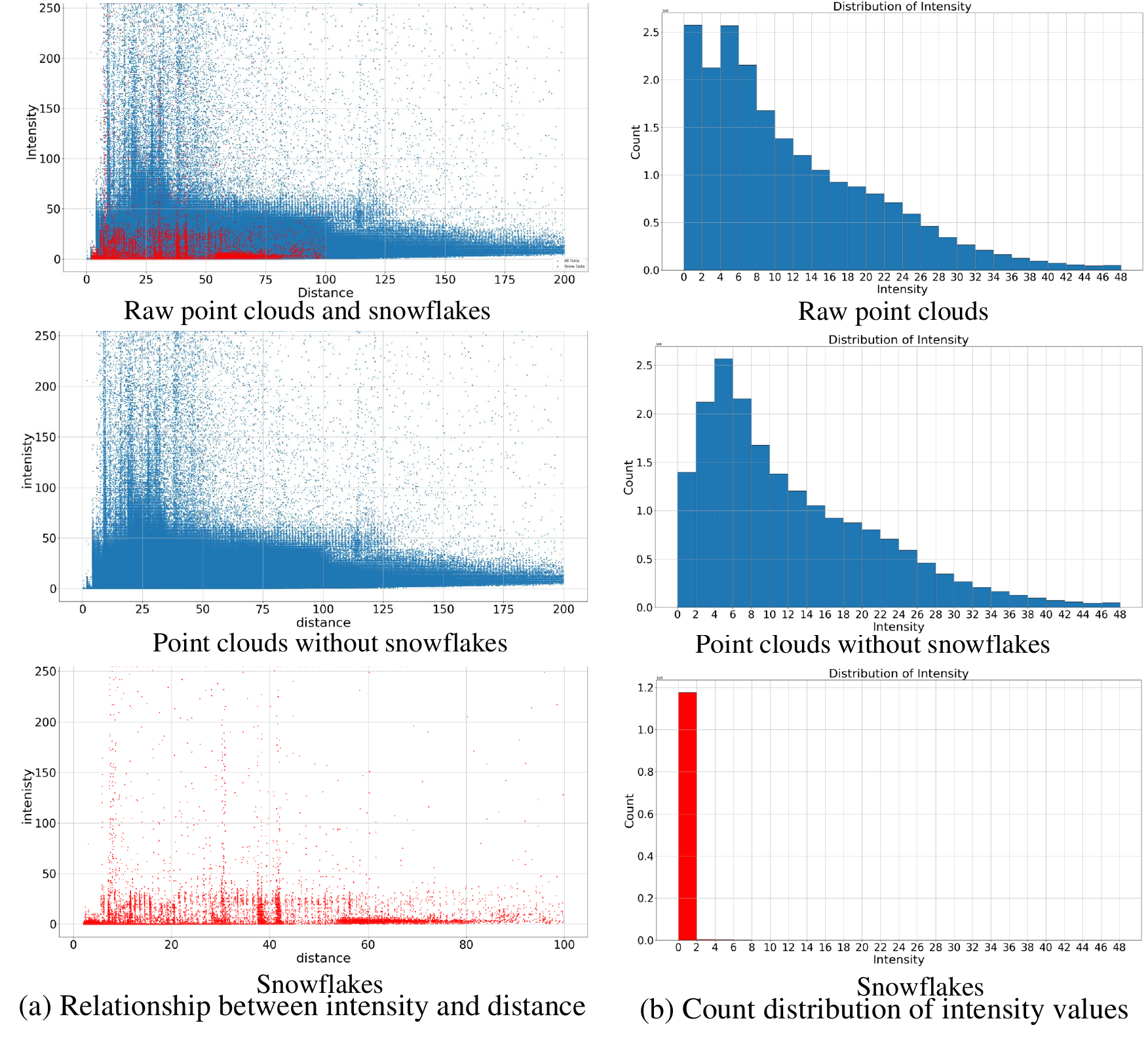} 
\caption{\textbf{Intensity Distribution of Snowy Point Clouds}. Using WADS dataset as an example, 100 frames were randomly selected for analysis. The snowflake points exhibit significantly low intensity values, primarily ranging from 0 to 2.}
\label{statistical}
\vspace{-0.2cm}
\end{figure}

\begin{figure} [t] 
\centering  
\includegraphics[width=0.95\linewidth]{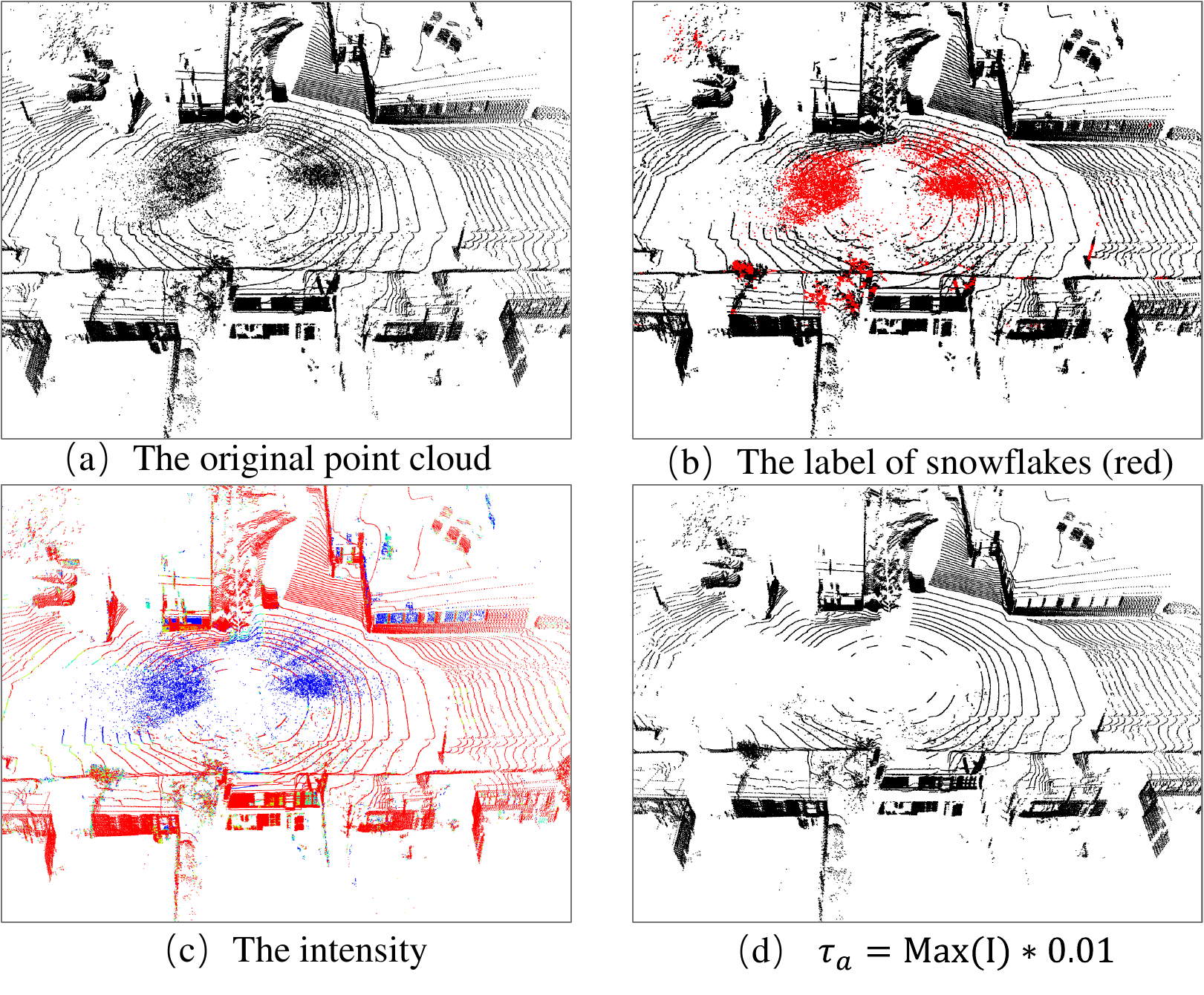}  
\caption{\textbf{Visualization of Snow Filtering Results using Intensity Threshold.} (a) The original unprocessed point cloud. (b) Ground-truth labels indicating snowflake points. (c) The intensity value visualization of the original point cloud, where color-coding ranges from red (highest intensity) to blue (lowest intensity).
(d) The filtered result after applying the intensity threshold of 
$\tau_a = \text{Max}(I) * 0.01$.}  
\label{snow_filter_result}  
\vspace{-0.5cm}
\end{figure}

 \begin{figure}[t]
\centering
\includegraphics[width=0.48\textwidth]{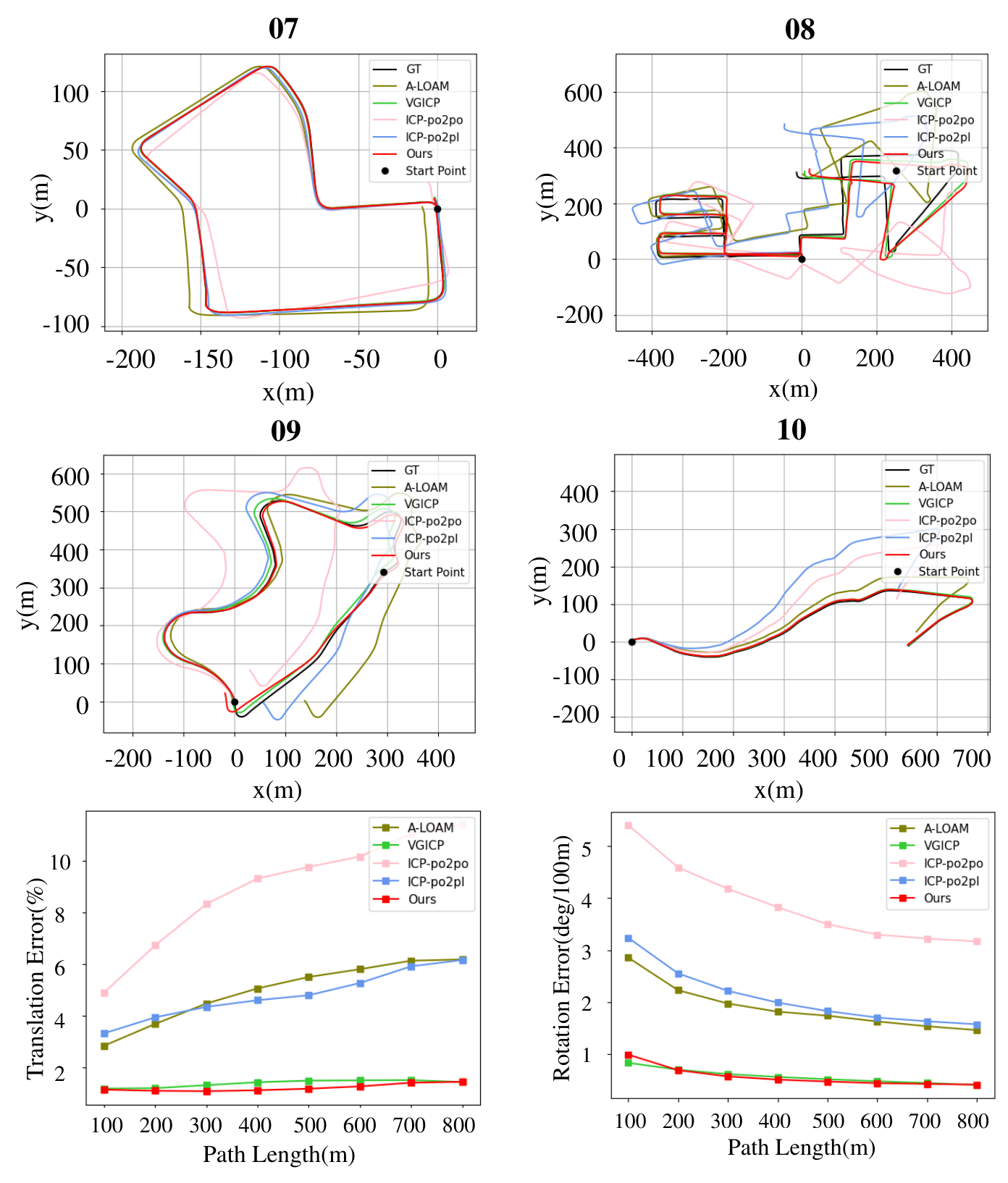} 
\caption{Visualization of the trajectories and the average rotation error and translation error of KITTI on sequences 07-10.}
\label{kitti_traj_error}
\vspace{-0.3cm}
\end{figure}

 \begin{figure}[t]
\centering
\includegraphics[width=0.48\textwidth]{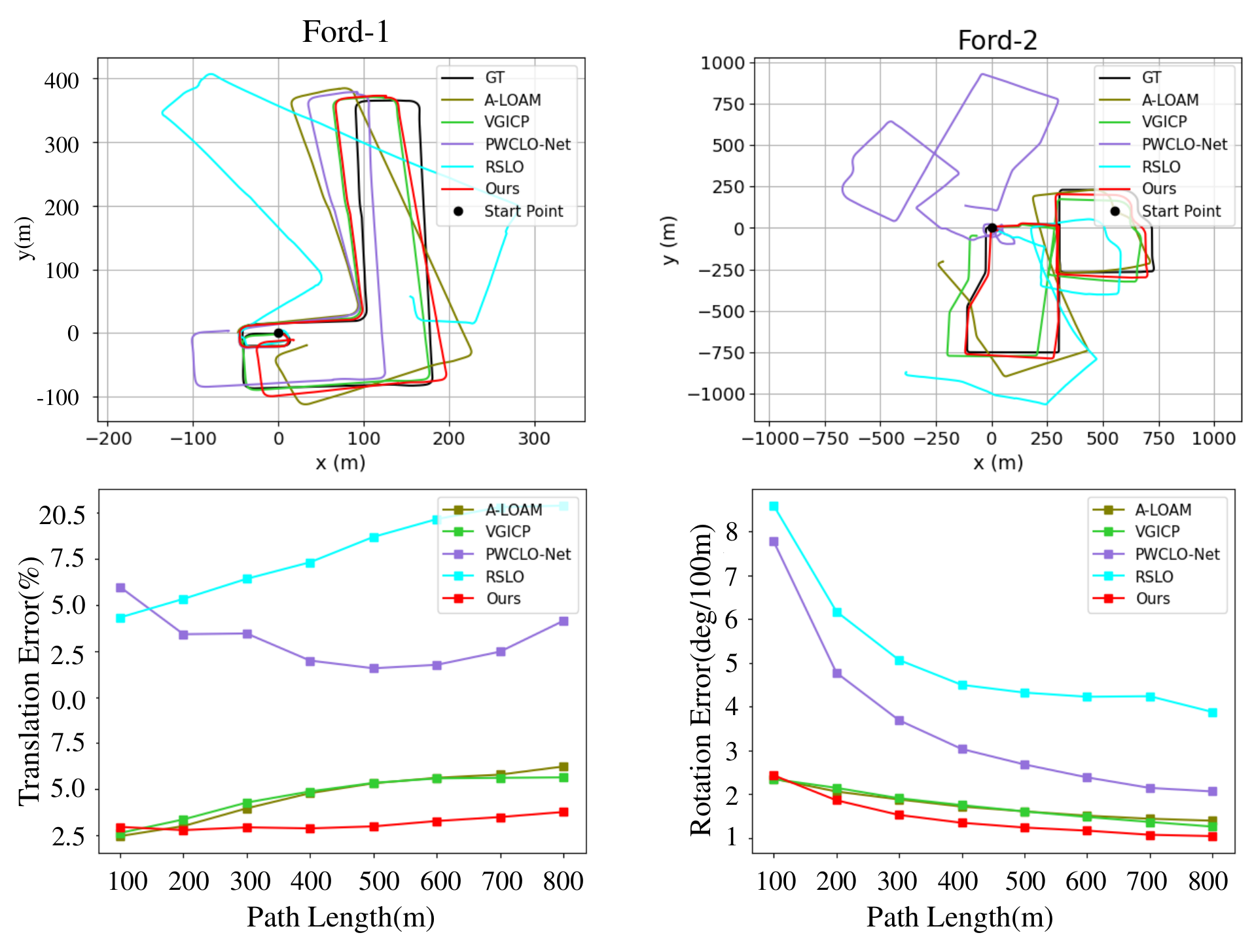} 
\caption{Visualization of the trajectories and the average rotation error and translation error on Ford sequences.}
\label{ford_traj_error}
\vspace{-0.2cm}
\end{figure}

 \begin{figure}[t]
\centering
\includegraphics[width=0.48\textwidth]{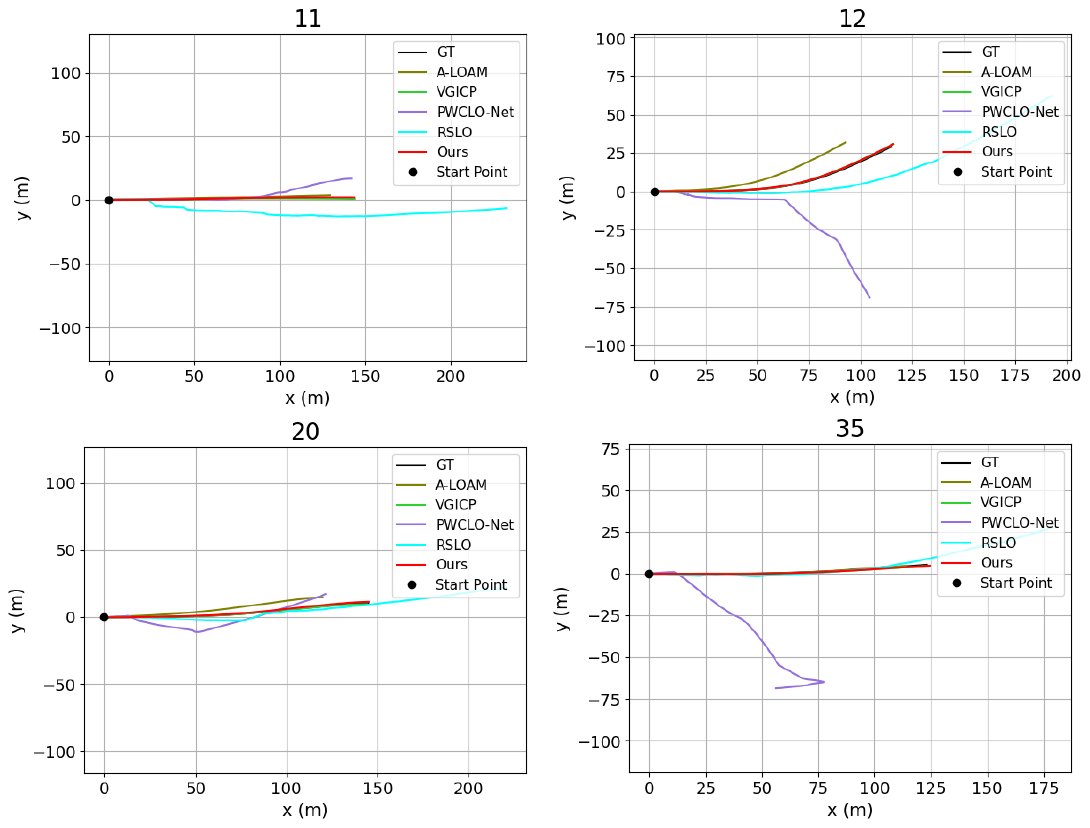} 
\caption{Visualization of the trajectories and the average rotation error and translation error of sequences 11,12,20,35 on WADS.}
\label{traj_error_wads}
\end{figure}

\subsection{visualization of intensity threshold mask}

First, we analyzed the intensity value distribution of snowflake points across the entire scene under snowfall conditions, as illustrated in~\Cref{statistical}. The intensity values of snowflake points are clearly concentrated in the lower range, distinguishing them from the overall point cloud distribution. 
In our approach, we segregate snowflake points from the snowy dataset with the intensity threshold mask, we employ the intensity threshold mask on diverse snowfall datasets and observe a high success rate in the elimination of the majority of snowflake points.
 The results of this thresholding process are visualized~\Cref{snow_filter_result}. 
A small number of residual snowflake points, which are not captured by the threshold, remain sparsely and discretely distributed within the point cloud. These remaining points can be further identified and eliminated as typical noise using the PSM and MPWP modules. This systematic approach ensures the comprehensive removal of snowflake points, thereby improving the efficiency of subsequent odometry computations.

\vspace{-0.2cm}
\subsection{Visualization of PSM and PPWP} 
Visualization examples for PSM and PPWP (including MPWP and $M$) are presented in Figure~\ref{fig:wads_weight_viz}. The first row illustrates the outputs of each module under clear weather conditions, while the second row displays their performance in snowy weather. As shown in panel (a), the point cloud is segmented into multiple patches based on superpoints. 
Panel (b) demonstrates that PSM assigns lower scores to sparse points (where accurate correspondences are difficult to establish) and scattered snowflake patches, while giving higher scores to dense clusters.
In particular, in snowy scenarios, PSM tends to give high scores to dense snowflake clusters near the sensor. Panel (c) shows that MPWP adaptively assigns lower confidence values to most snowflake points, effectively compensating for the limitations of PSM. After applying the threshold mask, as depicted in panel (d), nearly all snowflake points are filtered out. The few remaining points are assigned minimal weights by both PSM and MPWP, which significantly enhances pose estimation accuracy.  These findings highlight the crucial roles of the PSM, MPWP, and threshold mask modules in mitigating the negative impact of snowflake noise on LiDAR odometry, thereby ensuring robust performance in snowy environments.

\vspace{-0.2cm}
\subsection{Visualization of trajectories and errors}
\paragraph{Visualization on KITTI Odoemtry}
~\Cref{kitti_traj_error} illustrates the trajectory plots of our proposed method in comparison with selected state-of-the-art methods, namely A-LOAM, VGICP, ICP-po2po and ICP-po2pl, across sequences 07-10 of the KITTI Odometry dataset. Our method demonstrates a remarkable alignment with the ground truth (GT), particularly excelling in achieving precise loop closures on sequences 07 and 09. This accomplishment is especially notable given that our approach operates without the support of back-end optimization.
Moreover, ~\Cref{kitti_traj_error} presents a visualization of the average translational and rotational errors across all possible subsequences for sequences 07-10. Our method consistently exhibits the lowest error rates as the path length extends. In contrast, A-LOAM, ICP-po2pl, and ICP-po2pl, which lack back-end optimization, exhibit increasing average translation errors and consistently high rotation errors as path lengths grow. 
Although VGICP demonstrates more stable performance compared to other methods, its errors still slightly exceed those of our approach.

\paragraph{Visualization on Ford} ~\Cref{ford_traj_error} illustrates the trajectory visualization results for Ford-1 and Ford-2, comparing the trajectories of A-LOAM, VGICP, PWCLO-Net, RSLO, and our method ('Ours') with the ground-truth trajectory ('GT'). Our trajectory aligns best with the GT in both sequences. Notably, PWCLO-Net incorrectly estimates the relative poses during the initial stationary phase of Ford-2, resulting in significant deviation from the ground-truth trajectory. Even without ground-truth pose supervision, our method effectively handles such scenarios. In the last row, \Cref{ford_traj_error} shows the average translation and rotation errors across the subsequences, demonstrating that our method consistently achieves the lowest error range. These experimental phenomena show that we can have good generalization even in dynamic environments with similar scenes.

\paragraph{Visualization on WADS}   
The trajectory results for several WADS sequences are presented in \Cref{traj_error_wads}, where the trajectories of A-LOAM, VGICP, PWCLO-Net, RSLO, and our method ("Ours") are compared against the ground-truth trajectory ("GT"). We did not report the mean trajectory errors due to the short distances covered in the WADS dataset. It can be observed that traditional methods achieve good results, while deep learning-based LiDAR odometry methods (PWCLO-Net and RSLO) perform poorly. PWCLO-Net, in particular, frequently exhibits abrupt trajectory changes caused by errors in relative pose estimation for certain frames, leading to significant deviations in absolute pose orientation.      
Our method demonstrates superior generalization in snowy conditions compared to other approaches, owing to the effectiveness of its specialized modules.

\vspace{-0.2cm}
\section{Runtime}
\begin{table}[t]	
    \caption{Comparison of the overall runtime for LO methods.}
\centering
\resizebox{1\linewidth}{!}{
    \begin{tabular}{c||c|c|c|c|c|c|c} \toprule[1.5pt]
    {Methods}  &ICP-po2po  &ICP-po2pl &VGICP &PWCLO-Net &RSLO &HPPLO-Net &Ours \\ \hline
    {Time(ms)}  &81.6 &87.6 &94.2 &21.8&64.2 &27.8 &\textbf{17.2} \\
        \toprule[1.5pt]
\end{tabular}}
\label{tab:time}
\vspace{-0.3cm}	
\end{table}
 \begin{table}[t]	
    \caption{Comparison of the snow removal runtime and pose prediction error.}
\centering
\resizebox{0.75\linewidth}{!}{
    \begin{tabular}{c||c|c|c|c} \toprule[1.5pt]
    {Methods}  &DROR  &DSOR &LiSnowNet   &Ours \\ \hline
    {Time(ms)} &369 &510  &5.6 &\textbf{1.48} \\        \hline
    {Avg\_error\_R} &2.00 &2.14  &2.57 &\textbf{1.84} \\
    {Avg\_error\_t}  &1.97 &1.06  &1.54 &\textbf{1.15} \\
    \toprule[1.5pt]
\end{tabular}}
\label{tab:snow_time}
\vspace{-0.3cm}	
\end{table}

 \paragraph{Runtime Comparison of Odometry Methods} We evaluated the runtime of our method in comparison to several others, including ICP-po2po, ICP-po2pl, VGICP, PWCLO-Net, RSLO, and HPPLO-Net. The evaluation results are detailed in Table~\ref{tab:time}, collected under identical hardware conditions. These results represent the average pose estimation time for 100 point cloud pairs and exclude the time spent on data preprocessing.          
The traditional methods ICP-po2po, ICP-po2pl, and VGICP are generally time-consuming due to their inherent characteristics.
Notably, our method demonstrated the shortest runtime among the compared methods. Compared to HPPLO-Net, we removed its complex multi-layer normal estimation module and optimized various implementation details, which greatly improved the runtime efficiency on our new hardware platform and enabled real-time performance.

 \paragraph{Runtime Comparison of Snow Removal Methods}

 We replaced the threshold mask module with several widely-used denoising methods and compared their pose prediction accuracy and runtime (see Table~\ref{tab:snow_time} for details). The experimental results indicate that different denoising methods have a relatively minor impact on the final pose estimation results. This is primarily because pose estimation algorithms inherently require only a sufficient number of accurate matched point pairs to solve for a precise pose transformation matrix. Furthermore, other modules within our model further enhance the robustness and accuracy of point pair matching, making the final results less sensitive to the omission or misclassification of a small number of noise points. Considering the strict real-time requirements of autonomous driving, our method, which efficiently removes a large number of noise points, is highly suitable for pose estimation tasks. In contrast, other methods, due to their high computational complexity and longer processing times, are not the optimal choice for such applications.

 \vspace{-0.2cm}
\section{Conclusion}
\label{sec:conclusion}

In conclusion, we have introduced a novel unsupervised LiDAR odometry framework that generalizes effectively under normal and snowy weather conditions without requiring manual labeling. The proposed method integrates a Patch Spatial Measure for scene sparsity evaluation, a Patch Point Weight Predictor with a statistical threshold mask for quickly clustered snowflake noise removal and point-wise confidence assessment. 
This integrated approach enables real-time denoising while maintaining high pose prediction accuracy by focusing on robust regions.
Comprehensive evaluations on KITTI Odometry, Ford, and WADS datasets demonstrate the framework's robustness and efficiency across various weather conditions. The results highlight its potential to enhance autonomous navigation systems in real-world scenarios with variable weather. 
This work underscores the feasibility of unsupervised methods for achieving reliable and efficient odometry in challenging environments.
While the current implementation shows promising results, future work will extend the model's applicability to additional challenging conditions like heavy rain and fog. These developments will contribute to more resilient autonomous navigation systems capable of handling diverse environmental conditions, representing a significant advancement in real-world LiDAR odometry applications.

\section*{Acknowledgments}
This work was supported by the Science and Technology Development Fund of Macau SAR (File No. AGJ-2021-0046, AFJ-2022-0123, AFJ-2023-0067, SKL-IOTSC(UM)-2021-2023), and the startup project of Macau University (SRG2021-00022-IOTSC).

\bibliography{main}

\begin{thebibliography}{10}
\providecommand{\url}[1]{#1}
\csname url@samestyle\endcsname
\providecommand{\newblock}{\relax}
\providecommand{\bibinfo}[2]{#2}
\providecommand{\BIBentrySTDinterwordspacing}{\spaceskip=0pt\relax}
\providecommand{\BIBentryALTinterwordstretchfactor}{4}
\providecommand{\BIBentryALTinterwordspacing}{\spaceskip=\fontdimen2\font plus
\BIBentryALTinterwordstretchfactor\fontdimen3\font minus \fontdimen4\font\relax}
\providecommand{\BIBforeignlanguage}[2]{{%
\expandafter\ifx\csname l@#1\endcsname\relax
\typeout{** WARNING: IEEEtran.bst: No hyphenation pattern has been}%
\typeout{** loaded for the language `#1'. Using the pattern for}%
\typeout{** the default language instead.}%
\else
\language=\csname l@#1\endcsname
\fi
#2}}
\providecommand{\BIBdecl}{\relax}
\BIBdecl

\bibitem{geiger2013vision}
A.~Geiger, P.~Lenz, C.~Stiller, and R.~Urtasun, ``Vision meets robotics: The kitti dataset,'' \emph{The International Journal of Robotics Research}, vol.~32, no.~11, pp. 1231--1237, 2013.

\bibitem{zhang2014loam}
J.~Zhang and S.~Singh, ``Loam: Lidar odometry and mapping in real-time,'' in \emph{Robotics: Science and systems}, 2014, pp. 1--9.

\bibitem{chen2015deepdriving}
C.~Chen, A.~Seff, A.~Kornhauser, and J.~Xiao, ``Deepdriving: Learning affordance for direct perception in autonomous driving,'' in \emph{Proceedings of the IEEE international conference on computer vision}, 2015, pp. 2722--2730.

\bibitem{zang2019impact}
S.~Zang, M.~Ding, D.~Smith, P.~Tyler, T.~Rakotoarivelo, and M.~A. Kaafar, ``The impact of adverse weather conditions on autonomous vehicles: How rain, snow, fog, and hail affect the performance of a self-driving car,'' \emph{IEEE vehicular technology magazine}, vol.~14, no.~2, pp. 103--111, 2019.

\bibitem{carrilho2018statistical}
A.~Carrilho, M.~Galo, and R.~Santos, ``Statistical outlier detection method for airborne lidar data,'' \emph{The International Archives of the Photogrammetry, Remote Sensing and Spatial Information Sciences}, pp. 87--92, 2018.

\bibitem{charron2018noising}
N.~Charron, S.~Phillips, and S.~L. Waslander, ``De-noising of lidar point clouds corrupted by snowfall,'' in \emph{Conference on Computer and Robot Vision}.\hskip 1em plus 0.5em minus 0.4em\relax IEEE, 2018, pp. 254--261.

\bibitem{kurup2021dsor}
A.~Kurup and J.~Bos, ``Dsor: A scalable statistical filter for removing falling snow from lidar point clouds in severe winter weather,'' \emph{arXiv preprint arXiv:2109.07078}, 2021.

\bibitem{roriz2021dior}
R.~Roriz, A.~Campos, S.~Pinto, and T.~Gomes, ``Dior: A hardware-assisted weather denoising solution for lidar point clouds,'' \emph{IEEE Sensors Journal}, pp. 1621--1628, 2021.

\bibitem{heinzler2020cnn}
R.~Heinzler, F.~Piewak, P.~Schindler, and W.~Stork, ``Cnn-based lidar point cloud de-noising in adverse weather,'' \emph{IEEE Robotics and Automation Letters}, pp. 2514--2521, 2020.

\bibitem{seppanen20224denoisenet}
A.~Sepp{\"a}nen, R.~Ojala, and K.~Tammi, ``4denoisenet: Adverse weather denoising from adjacent point clouds,'' \emph{IEEE Robotics and Automation Letters}, pp. 456--463, 2022.

\bibitem{yu2022lisnownet}
M.-Y. Yu, R.~Vasudevan, and M.~Johnson-Roberson, ``Lisnownet: Real-time snow removal for lidar point clouds,'' in \emph{IEEE/RSJ International Conference on Intelligent Robots and Systems}, 2022, pp. 6820--6826.

\bibitem{velas2018cnn}
M.~Velas, M.~Spanel, M.~Hradis, and A.~Herout, ``Cnn for imu assisted odometry estimation using velodyne lidar,'' in \emph{IEEE International Conference on Autonomous Robot Systems and Competitions}, 2018, pp. 71--77.

\bibitem{li2019net}
Q.~Li, S.~Chen, C.~Wang, X.~Li, C.~Wen, M.~Cheng, and J.~Li, ``Lo-net: Deep real-time lidar odometry,'' in \emph{Proceedings of the IEEE/CVF Conference on Computer Vision and Pattern Recognition}, 2019, pp. 8473--8482.

\bibitem{wang2019deeppco}
W.~Wang, M.~R.~U. Saputra, P.~Zhao, P.~Gusmao, B.~Yang, C.~Chen, A.~Markham, and N.~Trigoni, ``Deeppco: End-to-end point cloud odometry through deep parallel neural network,'' in \emph{IEEE/RSJ International Conference on Intelligent Robots and Systems}, 2019, pp. 3248--3254.

\bibitem{translo}
N.~Charron, S.~Phillips, and S.~L. Waslander, ``Translo: A window-based masked point transformer framework for large-scale lidar odometry,'' in \emph{Proceedings of the AAAI Conference on Artificial Intelligence}, 2023.

\bibitem{wang2021pwclo}
G.~Wang, X.~Wu, Z.~Liu, and H.~Wang, ``Pwclo-net: Deep lidar odometry in 3d point clouds using hierarchical embedding mask optimization,'' in \emph{Proceedings of the IEEE/CVF Conference on Computer Vision and Pattern Recognition}, 2021, pp. 15\,910--15\,919.

\bibitem{wang2022efficient}
G.~Wang, X.~Wu, S.~Jiang, Z.~Liu, and H.~Wang, ``Efficient 3d deep lidar odometry,'' \emph{IEEE transactions on pattern analysis and machine intelligence}, 2022.

\bibitem{Ali_2023_ICCV}
S.~A. Ali, D.~Aouada, G.~Reis, and D.~Stricker, ``Delo: Deep evidential lidar odometry using partial optimal transport,'' in \emph{Proceedings of the IEEE/CVF International Conference on Computer Vision (ICCV) Workshops}, 2023.

\bibitem{cho2020unsupervised}
Y.~Cho, G.~Kim, and A.~Kim, ``Unsupervised geometry-aware deep lidar odometry,'' in \emph{IEEE International Conference on Robotics and Automati}, 2020, pp. 2145--2152.

\bibitem{low2004Linear}
K.~L. Low, ``Linear least-squares optimization for point-to-plane icp surface registration,'' \emph{Chapel Hill}, 2004.

\bibitem{SelfVoxeLO}
Y.~Xu, Z.~Huang, K.-Y. Lin, X.~Zhu, J.~Shi, H.~Bao, G.~Zhang, and H.~Li, ``Selfvoxelo: Self-supervised lidar odometry with voxel-based deep neural networks,'' in \emph{Conference on Robot Learning}, 2021, pp. 115--125.

\bibitem{zhou2023hpplo}
B.~Zhou, Y.~Tu, Z.~Jin, C.~Xu, and H.~Kong, ``Hpplo-net: Unsupervised lidar odometry using a hierarchical point-to-plane solver,'' \emph{IEEE Transactions on Intelligent Vehicles}, pp. 2727--2739, 2023.

\bibitem{li20233d}
C.~Li, F.~Yan, S.~Wang, and Y.~Zhuang, ``A 3d lidar odometry for ugvs using coarse-to-fine deep scene flow estimation,'' \emph{Transactions of the Institute of Measurement and Control}, pp. 274--286, 2023.

\bibitem{park2020fast}
J.-I. Park, J.~Park, and K.-S. Kim, ``Fast and accurate desnowing algorithm for lidar point clouds,'' \emph{IEEE Access}, pp. 160\,202--160\,212, 2020.

\bibitem{wang2022scalable}
W.~Wang, X.~You, L.~Chen, J.~Tian, F.~Tang, and L.~Zhang, ``A scalable and accurate de-snowing algorithm for lidar point clouds in winter,'' \emph{Remote Sensing}, p. 1468, 2022.

\bibitem{qi2017pointnet}
C.~R. Qi, H.~Su, K.~Mo, and L.~J. Guibas, ``Pointnet: Deep learning on point sets for 3d classification and segmentation,'' in \emph{Proceedings of the IEEE Conference on Computer Vision and Pattern Recognition}, 2017, pp. 652--660.

\bibitem{rakotosaona2020pointcleannet}
M.-J. Rakotosaona, V.~La~Barbera, P.~Guerrero, N.~J. Mitra, and M.~Ovsjanikov, ``Pointcleannet: Learning to denoise and remove outliers from dense point clouds,'' in \emph{Computer Graphics Forum}, 2020, pp. 185--203.

\bibitem{bae2022slide}
G.~Bae, B.~Kim, S.~Ahn, J.~Min, and I.~Shim, ``Slide: Self-supervised lidar de-snowing through reconstruction difficulty,'' in \emph{Proceedings of the European Conference on Computer Vision}, 2022, pp. 283--300.

\bibitem{qi2017pointnet++}
C.~R. Qi, L.~Yi, H.~Su, and L.~J. Guibas, ``Pointnet++: Deep hierarchical feature learning on point sets in a metric space,'' in \emph{Advances in Neural Information Processing Systems}, 2017.

\bibitem{wu2019pointconv}
W.~Wu, Z.~Qi, and L.~Fuxin, ``Pointconv: Deep convolutional networks on 3d point clouds,'' in \emph{Proceedings of the IEEE/CVF Conference on Computer Vision and Pattern Recognition}, 2019, pp. 9621--9630.

\bibitem{zhang2023detecting}
C.~Zhang, J.~Han, Y.~Zou, K.~Dong, Y.~Li, J.~Ding, and X.~Han, ``Detecting the anomalies in lidar pointcloud,'' \emph{arXiv preprint arXiv:2308.00187}, 2023.

\bibitem{moran1950notes}
P.~A. Moran, ``Notes on continuous stochastic phenomena,'' \emph{Biometrika}, pp. 17--23, 1950.

\bibitem{li2007beyond}
H.~Li, C.~A. Calder, and N.~Cressie, ``Beyond moran's i: testing for spatial dependence based on the spatial autoregressive model,'' \emph{Geographical analysis}, pp. 357--375, 2007.

\bibitem{wu2020pointpwc}
W.~Wu, Z.~Y. Wang, Z.~Li, W.~Liu, and L.~Fuxin, ``Pointpwc-net: Cost volume on point clouds for (self-) supervised scene flow estimation,'' in \emph{Proceedings of the European Conference on Computer Vision}, 2020, pp. 88--107.

\bibitem{geiger2012we}
A.~Geiger, P.~Lenz, and R.~Urtasun, ``Are we ready for autonomous driving? the kitti vision benchmark suite,'' in \emph{Proceedings of the IEEE/CVF Conference on Computer Vision and Pattern Recognition}, 2012, pp. 3354--3361.

\bibitem{aloam}
T.~Qin, S.~Cao, J.~Behley, and C.~Stachniss, ``Advanced implementation of loam,'' in \emph{https://github.com/HKUST-Aerial-Robotics/A-LOAM}, 2019.

\bibitem{ICP}
J.~Giseop~Kim, ``Pyicp slam.'' in \emph{\url{https://github.com/JustWon/PyICP-SLAM}}, 2023.

\bibitem{koide2021voxelized}
K.~Koide, M.~Yokozuka, S.~Oishi, and A.~Banno, ``Voxelized gicp for fast and accurate 3d point cloud registration,'' in \emph{IEEE International Conference on Robotics and Automation}, 2021, pp. 11\,054--11\,059.

\bibitem{velas2016collar}
M.~Velas, M.~Spanel, and A.~Herout, ``Collar line segments for fast odometry estimation from velodyne point clouds,'' in \emph{IEEE International Conference on Robotics and Automation}, 2016, pp. 4486--4495.

\bibitem{ali2023delo}
S.~A. Ali, D.~Aouada, G.~Reis, and D.~Stricker, ``Delo: Deep evidential lidar odometry using partial optimal transport,'' in \emph{Proceedings of the IEEE/CVF International Conference on Computer Vision}, 2023, pp. 4517--4526.

\bibitem{nubert2021self}
J.~Nubert, S.~Khattak, and M.~Hutter, ``Self-supervised learning of lidar odometry for robotic applications,'' in \emph{IEEE International Conference on Robotics and Automation}, 2021, pp. 9601--9607.

\bibitem{xu2022robust}
Y.~Xu, J.~Lin, J.~Shi, G.~Zhang, X.~Wang, and H.~Li, ``Robust self-supervised lidar odometry via representative structure discovery and 3d inherent error modeling,'' \emph{IEEE Robotics and Automation Letters}, pp. 1651--1658, 2022.

\end{thebibliography}

\begin{IEEEbiography}[{\includegraphics[width=1in,height=1.25in,keepaspectratio]{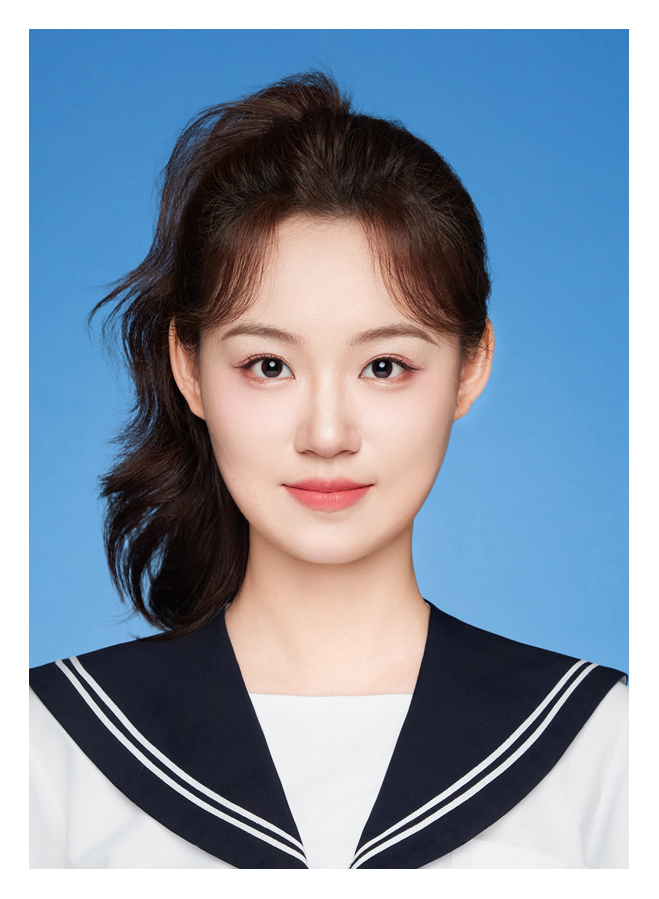}}]
 {Beibei Zhou} received her Ph.D. degree at the PCA Lab, he Key Lab of Intelligent Perception
and Systems for High-Dimensional Information of
Ministry of Education, School of Computer Science and Engineering, Nanjing University of Science and Technology, Nanjing, China, in 2024. She is currently working in Shanghai Polytechnic University. From 2023 to 2024, she was a visiting scholar with the State Key Laboratory of Internet of Things for Smart City (SKL-IOTSC), University of Macau, Macau.
 Her research interests include visual/Lidar odometry, 3D scene flow estimation, and computer vision.
\end{IEEEbiography} 
\vspace{-0.3cm}

\begin{IEEEbiography}[{\includegraphics[width=1in,height=1.25in,keepaspectratio]{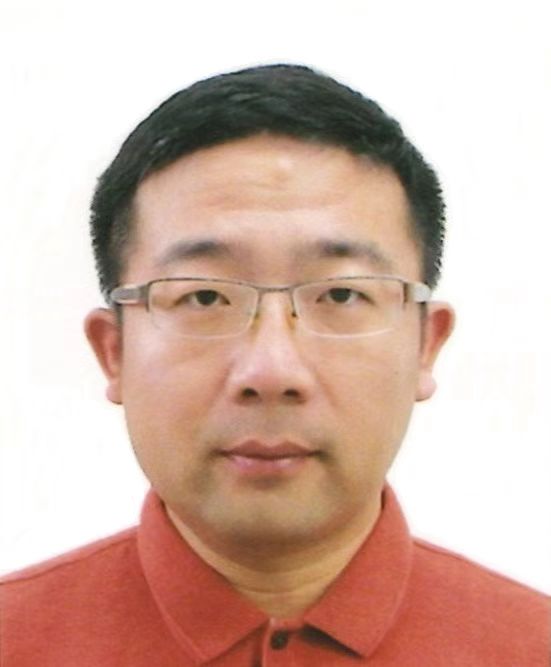}}]
 {Zhiyuan Zhang} received the Ph.D. degree in computer science from the National University of Singapore in 2015. He is currently an Assistant Professor at the School of Computing and Information Systems, Singapore Management University. Previously, he was an Assistant Professor at the Ningbo Innovation Center, Zhejiang University, and a Research Fellow at DMand Lab, Singapore University of Technology and Design. From 2014 to 2017, he worked as a Staff Researcher at Lenovo, leading computer vision projects. His research interests include 3D point cloud processing, lightweight deep learning, and 3D biometrics.
\end{IEEEbiography} 
\vspace{-0.3cm}


\begin{IEEEbiography}[{\includegraphics[width=1in,height=1.25in,keepaspectratio]{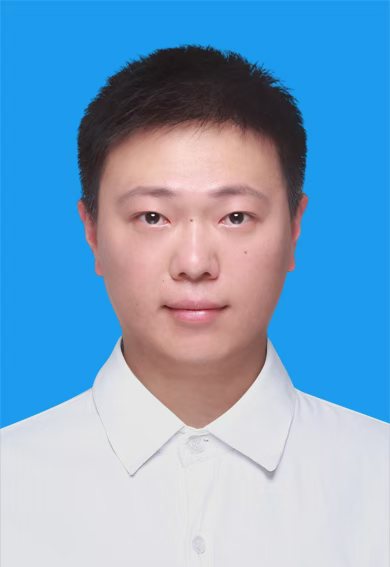}}]
 {Zhenbo Song} (Member, IEEE) received the Ph.D. degree in Control science and engineering from Nanjing University of Science and Technology, China, in 2022, where he is currently an Associate Professor. From 2018 to 2019, he was a Visiting Scholar with the Australian National University, Canberra, Australia. His current research interests include autonomous driving, image processing, 3-D reconstruction, and deep learning.
\end{IEEEbiography} 
\vspace{-0.3cm}

\begin{IEEEbiography}[{\includegraphics[width=1in,height=1.25in,keepaspectratio]{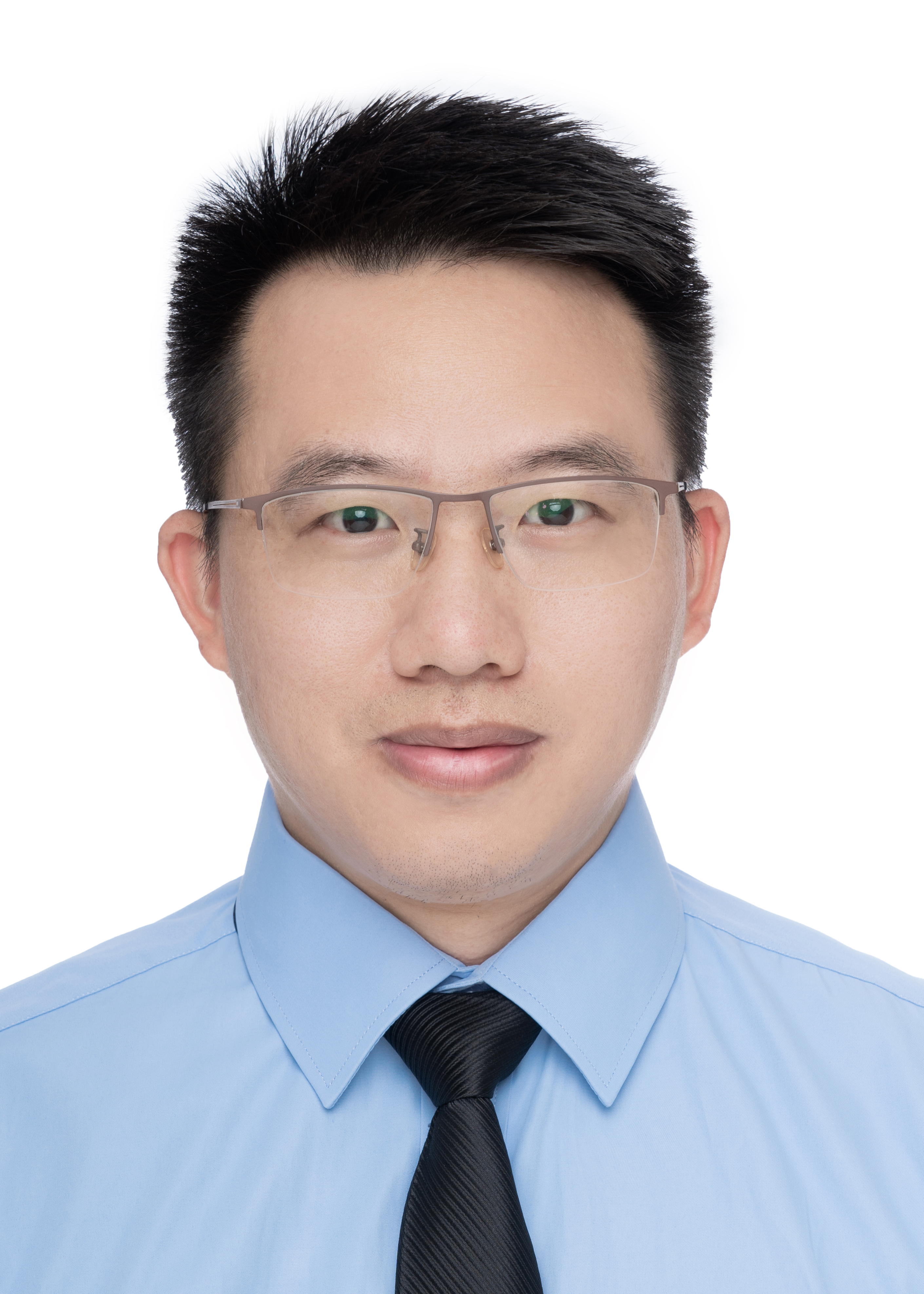}}]
 {Jianhui Guo} received the BS degree, MS degree and PHD degree from the Nanjing University of Science and Technology, Nanjing, China, in 2003, 2005 and 2008, respectively. In 2008, he joined the Nanjing Institute of Electronics Technology as a senior engineer.  He is now an associate professor in the School of Computer Science and engineering of Nanjing University of Science and Technology. His research interests include machine learning, data mining, pattern recognition, and intelligent robot, and information fusion. 
\end{IEEEbiography} 
\vspace{-0.3cm}

 \begin{IEEEbiography}[{\includegraphics[width=1in,height=1.25in,keepaspectratio]{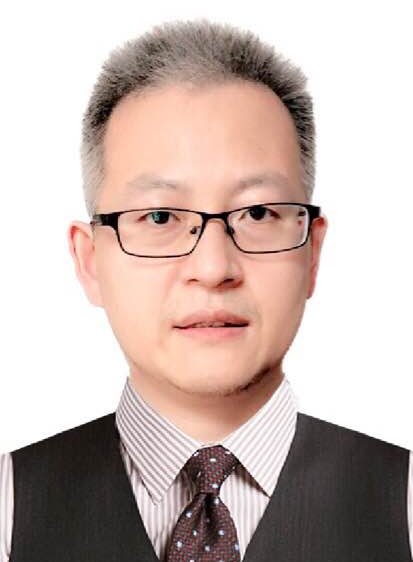}}]
 {Hui Kong}
received the Ph.D. degree in computer vision from Nanyang Technological University, Singapore, in 2007. He is currently an Associate Professor with the State Key Laboratory of Internet of Things for Smart City (SKL-IOTSC), Department of Electromechanical Engineering (EME), University of Macau, Macau. His research interests include sensing and perception for autonomous driving, SLAM, mobile robotics, multi-view geometry, and motion planning.
\end{IEEEbiography} 
\vspace{-0.3cm}

\bibliographystyle{IEEEtran}

\vfill

\end{document}